\documentclass[sn-mathphys-num]{sn-jnl}

\usepackage{graphicx}%
\usepackage{multirow}%
\usepackage{amsmath,amssymb,amsfonts}%
\usepackage[title]{appendix}%
\usepackage{xcolor}%
\usepackage{textcomp}%
\usepackage{manyfoot}%
\usepackage{booktabs}%
\usepackage{algorithm}%
\usepackage{algorithmicx}%
\usepackage{algpseudocode}%
\usepackage{listings}%

\usepackage{colortbl}
\usepackage{soul}
\usepackage{makecell}
\usepackage{capt-of}
\usepackage{lmodern}
\usepackage[T1]{fontenc}
\usepackage{afterpage}

\definecolor{customgreen}{rgb}{0.25, 0.69, 0.65}
\definecolor{customyellow}{rgb}{1, 1, 0.38}

\begin{document}

\title[HOI-Synth]{Leveraging Synthetic Data for Enhancing Egocentric Hand-Object Interaction Detection}

\author*[1,2]{\fnm{Rosario} \sur{Leonardi}}\email{rosario.leonardi@unict.it}
\author[1,2]{\fnm{Antonino} \sur{Furnari}}\email{antonino.furnari@unict.it}
\author[1,2]{\fnm{Francesco} \sur{Ragusa}}\email{francesco.ragusa@unict.it}
\author[1,2]{\fnm{Giovanni Maria} \sur{Farinella}}\email{giovanni.farinella@unict.it}

\affil*[1]{\orgdiv{Department of Mathematics and Computer Science}, \orgname{University of Catania}, \orgaddress{\city{Catania}, \country{Italy}}}
\affil[2]{\orgname{Next Vision s.r.l.}, \orgaddress{\city{Catania}, \country{Italy}}}


\abstract{In this work, we explore the role of synthetic data in improving the detection of Hand-Object Interactions from egocentric images. Through extensive experimentation and comparative analysis on \textit{VISOR}, \textit{EgoHOS}, and \textit{ENIGMA-51} datasets, our findings demonstrate the potential of synthetic data to significantly improve HOI detection, particularly when real labeled data are scarce or unavailable. By using synthetic data and only $10\%$ of the real labeled data, we achieve improvements in \textit{Overall AP} over models trained exclusively on real data, with gains of $+5.67\%$ on \textit{VISOR}, $+8.24\%$ on \textit{EgoHOS}, and $+11.69\%$ on \textit{ENIGMA-51}. Furthermore, we systematically study how aligning synthetic data to specific real-world benchmarks with respect to objects, grasps, and environments, showing that the effectiveness of synthetic data consistently improves with better synthetic-real alignment. As a result of this work, we release a new data generation pipeline and the new \textit{HOI-Synth} benchmark, which augments existing datasets with synthetic images of hand-object interaction. These data are automatically annotated with hand-object contact states, bounding boxes, and pixel-wise segmentation masks. All data, code, and tools for synthetic data generation are available at: \url{https://fpv-iplab.github.io/HOI-Synth/}.}

\keywords{Synthetic Data, Egocentric Hand-Object Interaction Detection, Domain Adaptation}


\maketitle

\section{Introduction}\label{sec:intro}

Understanding human-object interactions is a key challenge in computer vision, with several applications in fields like collaborative robotics~\cite{edsinger2007human, carfi2021hand}, industrial behaviour analysis~\cite{sener2022assembly101, Ragusa2021TheMD}, human-computer interaction~\cite{lv2022deep}, and healthcare~\cite{besari2023hand}. Prior research has explored human-object interactions in egocentric vision, focusing on tasks such as action recognition~\cite{Damen2021RESCALING}, object state-change detection~\cite{Grauman2021Ego4DAT}, and hand-object interactions forecasting~\cite{liu2022joint}. A line of research focuses on detecting manipulated objects, identifying hands, and recognizing contact between hands and objects~\cite{Shan2020UnderstandingHH, Ragusa2021TheMD, leonardi2022egocentric, VISOR2022}, known as \textit{Hand-Object Interaction (HOI) detection}. Despite advancements in model development supported by egocentric datasets like \textit{VISOR}~\cite{VISOR2022}, performance remains heavily reliant on domain-specific data~\cite{Ragusa2021TheMD, leonardi2024exploiting}. The complexity of acquiring hand-object interaction annotations represents a significant challenge, as the process of data collection and labeling is resource-intensive, both in time and cost.

Synthetic data has been widely used to reduce the reliance on large amounts of labeled real-world data in fields like embodied AI~\cite{kolve2017ai2,savva2019habitat,xia2020interactive} and autonomous driving~\cite{Dosovitskiy17,fabbri21iccv}. However, its exploitation in egocentric vision, particularly in hand-object interaction detection, remains underexplored. This is due to the inherent challenges in generating realistic images that capture the complexities of hands, objects, and their physical interactions from egocentric perspective. As a result, several key open questions still need to be addressed: \textit{1) How large is the gap between synthetic and real data? 2) What are its main causes? 3) How can it be minimized? 4) Can synthetic data fully replace real-world data? 5) Is it possible to leverage synthetic data when real-world data is unlabeled? 6) Can it improve performance when only a small amount of real-world labeled data is available? 7) What scale of synthetic data is required? 8) Does aligning synthetic data more closely with real-world objects and environments provide additional benefits?}

\begin{figure}[t]
    \centering
    \includegraphics[width=\linewidth]{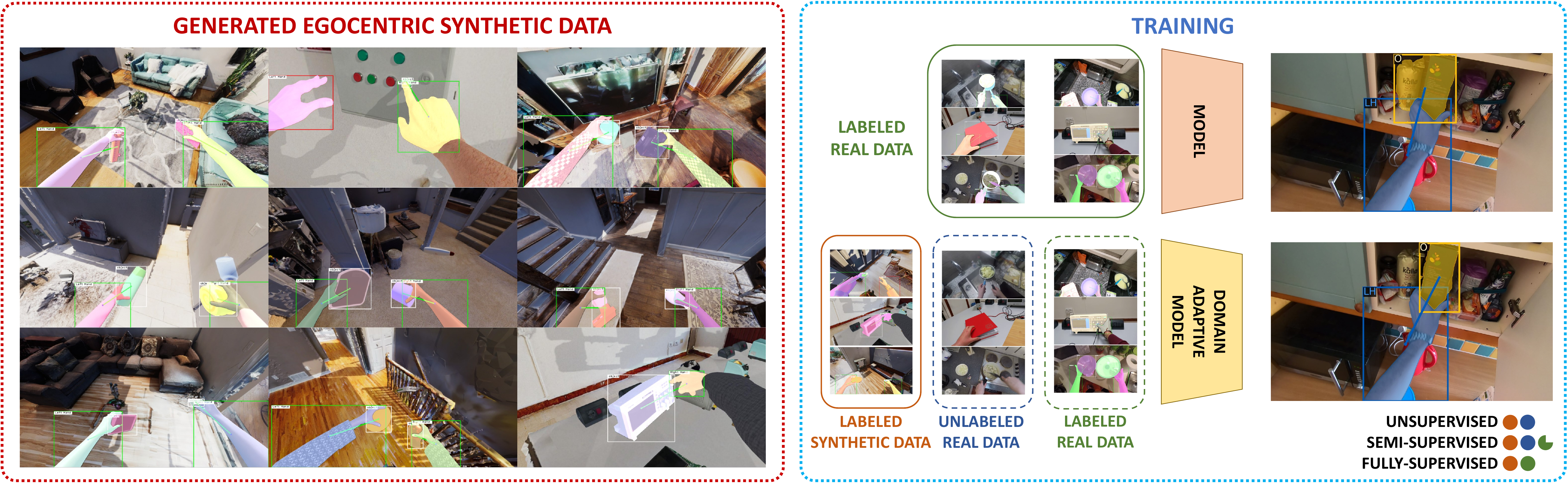}
    \caption{We explore the role of synthetic data in egocentric hand-object interaction detection by generating and automatically labeling synthetic datasets (left). We then analyze domain adaptation scenarios where models are trained using both synthetic and real unlabeled data, with varying amounts of labeled real data (right).}
    \label{fig:figure1}
\end{figure}

To further research in egocentric hand-object interaction detection and synthetic-to-real domain adaptation for egocentric vision, this work introduces a systematic investigation aimed at addressing the aforementioned questions. To enable this, we present a novel pipeline to generate realistic synthetic egocentric images of hand-object interactions across diverse environments, automatically labeled for the considered task (Figure~\ref{fig:figure1}-left). By exploiting our proposed pipeline, we generated three synthetic sets of egocentric data for three popular hand-object interaction detection benchmarks, i.e., \textit{VISOR}~\cite{VISOR2022}, \textit{EgoHOS}~\cite{EgoHos_jianbo_eccv22}, and \textit{ENIGMA-51}~\cite{ragusa2024enigma}. Together, the generated synthetic data and real egocentric datasets set a new benchmark, which we call \textit{HOI-Synth}. 

We study three different domain adaptation tasks: \textit{unsupervised domain adaptation}, where models are trained using synthetic data combined with unlabeled real data; \textit{semi-supervised domain adaptation}, which combines labeled synthetic data, unlabeled real data, and a limited amount of labeled real data; and \textit{fully supervised domain adaptation}, where models leverage both synthetic and labeled real data for training (Figure~\ref{fig:figure1}-right). We exploit \textit{HOI-Synth} to benchmark various approaches to domain adaptation for hand-object interaction detection, adapting prior works on domain adaptation for object detection~\cite{li2022cross,tarvainen2017mean,liu2021unbiased,ganin2015unsupervised}. 

We further conduct a detailed analysis targeting specific aspects of synthetic hand-object interactions to better align with real-world data, i.e.: environments, objects, and grasps. This analysis examines the impact of adapting each component individually and in combination, aiming to identify which aspects most significantly contribute to enhancing hand-object interaction detection.

Our findings offer key insights into the impact of synthetic data on model performance for egocentric hand-object interaction detection: A) A significant gap between synthetic and real data persists despite advancements in synthetic image generation, with models trained exclusively on synthetic data showing an AP of around 30\%-40\% compared to real-data-trained models. This gap is primarily due to limitations in image realism, grasp simulation, and environmental diversity. B) Applying domain adaptation techniques can significantly reduce the gap. In \textit{unsupervised domain adaptation}, models improve by 20\%-35\% AP when unlabeled real data are used. In \textit{semi-supervised adaptation}, models can achieve near real-data-trained performance using only 10\%-25\% of labeled real data. For \textit{fully-supervised domain adaptation}, combining labeled real and synthetic data leads to gains of 1\%-4\% AP. C) Most performance improvements are observed with synthetic data of around 10,000 images, but performance increases slightly as the dataset size increases. D) Using synthetic data aligned with a target domain, in terms of objects, hand poses and environments, improves performance in unsupervised domain adaptation. However, the advantage of domain-specific synthetic data diminishes when a small amount of real labeled data is used in semi-supervised adaptation.

This work extends our previous research~\cite{leonardi2025synthetic} by introducing new experiments and analyses aimed at improving both the generation and utility of synthetic hand-object interaction data. The main focus of this extension is the alignment of synthetic data with specific real-world target datasets. In particular, we explore how synthetic data can be better matched to real data by targeting three key aspects: objects, grasps, and environments. We then analyze how aligning each of these factors, individually or in combination, impacts model performance in domain adaptation scenarios. To guide this alignment, we leverage unlabeled real data together with pre-trained models such as \textit{DINOv2}~\cite{oquab2023dinov2} and \textit{MMPose}~\cite{mmpose2020} to identify the most relevant objects, hand poses, and environments for synthetic generation. Beyond alignment, we further expand the experimental study with additional ablations and comparisons, including experiments with a modern backbone (ConvNeXt-S) to assess architectural robustness, confirming that our findings generalize beyond the baseline network. We also provide a more detailed description of the simulator and data generation pipeline, covering camera configuration, rendering settings, clutter modeling, and automatic annotation. To ensure full reproducibility, we also include detailed implementation and computational metrics.

The key contributions of this work are: 1) A comprehensive investigation into egocentric hand-object interaction detection, evaluating the role of synthetic data in three domain adaptation contexts. Our study provides insights useful for future research on model design and experimental evaluation in this field. 2) The introduction of \textit{HOI-Synth}, a new benchmark designed to assess unsupervised, semi-supervised, and fully-supervised domain adaptation. This is the first benchmark to explore synthetic-to-real domain adaptation specifically for egocentric hand-object interaction detection. We also provide baseline results, showing the value of synthetic data and building a foundation for future research. 3) The development of a novel data generation pipeline and a simulator, which enable further research into the use of synthetic data for egocentric vision tasks. To encourage progress in the field, we publicly release the generated data, the simulator, and all code necessary to replicate our findings.


\section{Related Work}\label{sec:related_work}
 Our research builds on previous investigations in several areas relevant to the detection of hand-object interaction and the generation of synthetic data. The following subsections detail the main findings and contributions in these areas.

\subsection{Understanding Hand-Object Interaction Detection} 
The task of \textit{Hand-Object Interaction} detection was first formulated by the authors of~\cite{Shan2020UnderstandingHH} as a combination of detecting hands, determining hand contact states, and identifying manipulated objects. This task formulation was later extended in~\cite{cheng2023towards}, incorporating additional elements such as hand and object segmentation, secondary object detection, and prediction of grasp types. Other studies have addressed related problems, such as \textit{State Change Object Interaction Detection}~\cite{Grauman2021Ego4DAT} and \textit{Active Object Detection}~\cite{Fu2021SequentialDF}. In parallel, several studies have explored hand-object interactions in the context of egocentric vision. In the work of ~\cite{Ragusa2021TheMD}, the authors introduced the concept of \textit{Egocentric Human-Object Interaction} (EHOI) detection, which involves identifying manipulated objects and predicting interaction verbs. Additionally, egocentric hand-object interaction detection has been explored considering the industrial scenario~\cite{leonardi2022egocentric, leonardi2024exploiting, ragusa2024enigma} where the task is more challenging due to the presence of small objects and industrial tools. However, the diverse task formulations of these works have created challenges in comparing methods and tracking progress over time. 

Recently, the authors of~\cite{VISOR2022} introduced \textit{Hand-Object Segmentation} (HOS) as a standardized formulation for HOI detection. HOS involves estimating the contact relationships between hands and objects and segmenting both from a single RGB frame. Furthermore, this work introduced the \textit{VISOR} dataset~\cite{VISOR2022}, establishing a new benchmark for evaluating hand-object interactions in egocentric settings. 
The authors of~\cite{EgoHos_jianbo_eccv22} adopted a comparable formulation, focusing on segmenting hands and objects in egocentric scenarios. Their contribution also included the introduction of the \textit{EgoHOS} dataset, specifically designed to support research on egocentric hand-object interaction detection.

In this work, we adopt the HOS formulation and perform experiments on three datasets designed for hand-object interaction detection: \textit{VISOR}~\cite{VISOR2022}, \textit{EgoHOS}~\cite{EgoHos_jianbo_eccv22}, and \textit{ENIGMA-51}~\cite{ragusa2024enigma}. Through our analysis, we demonstrate how the use of synthetic data can lead to significant improvements in the performance of state-of-the-art hand-object interaction detectors.

\subsection{Synthetic Data Generation for Hand-Object Interaction Detection}
Synthetic data generation through simulators has become a crucial tool in computer vision, particularly for simulating real-world agents such as vehicles, robots, and pedestrians. Well-known simulators like \textit{CARLA}~\cite{Dosovitskiy17}, \textit{Gibson}~\cite{xiazamirhe2018gibsonenv, li2022igibson}, \textit{Habitat}~\cite{habitat19iccv, szot2021habitat}, and \textit{Isaac Sim}~\cite{isaacSim_nvidia_2021} have been extensively used to generate realistic scenarios for tasks such as autonomous driving, object detection, and robot navigation. Additionally, advancements in graphics and game engines, such as the \textit{RAGE}~\cite{rageEngine} engine used in \textit{Grand Theft Auto V}, have simplified the generation of synthetic datasets by providing highly detailed and customizable environments. These engines simulate realistic scenarios with rich urban landscapes and dynamic weather conditions, making them suitable for tasks like pedestrian tracking~\cite{fabbri21iccv, di2021learning}, as well as safety monitoring in construction sites~\cite{quattrocchi2023Outline_SAFER}. More recently, egocentric simulators, such as \textit{AI2Thor}~\cite{ai2thor}, have been developed to simulate human-object interactions, albeit with limitations. For example, \textit{AI2Thor} focuses on object manipulation actions without simulating human hands within the scene. \textit{EgoGen}~\cite{li2024egogen} is a simulator designed to generate realistic synthetic data for various egocentric vision tasks, including mapping, camera localization and tracking, and human mesh recovery from egocentric views. However, it is not specifically tailored to generate hand-object interaction data directly. Another example in the domain of egocentric vision is \textit{Aria Synthetic Environments}~\cite{avetisyan2024scenescript} (ASE), a simulator for generating realistic 3D data focused on tasks such as object detection and spatial understanding. ASE provides large-scale, procedurally generated indoor scenes with realistic object placement, ideal for 3D scene reconstruction. However, also in this case, it is not specifically designed for hand-object interaction tasks and does not model hand-object interactions directly. The work of~\cite{leonardi2024exploiting} introduces a simulator designed to generate synthetic data for hand-object interaction detection in industrial settings. The simulator generates RGB images, depth maps, and interaction labels, such as hand-object contact and pose annotations. However, its scope is limited to a specific industrial environment, which means it does not capture the full range of diverse interactions found in everyday scenarios. This makes it valuable for certain industrial applications but not readily adaptable to more generalized environments.

Despite the progress made in synthetic data generation for various domains, the application of these approaches to egocentric hand-object interaction detection remains largely underexplored. One of the main challenges lies in accurately replicating the dynamic nature of hand-object interactions in real environments. The complexity of modeling objects, human hands, and their interactions has made it difficult to generate realistic synthetic datasets. To overcome these limitations, we propose a new generation pipeline designed specifically for creating diverse and realistic egocentric images of hand-object interactions at large scales. Our pipeline exploits state-of-the-art developed modules to generate photorealistic and accurate hand-object interaction data.

\subsection{Domain Adaptation for Hand-Object Interaction Understanding}
In recent years, Domain Adaptation (DA) techniques have become a central focus in machine learning research~\cite{csurka2017domain, saito2018maximum, bousmalis2017unsupervised, zhuang2020comprehensive}. These approaches aim to mitigate the challenges posed by domain shifts, enabling models trained on one dataset (source domain) to perform well on a different dataset (target domain). Common methods for achieving this include adversarial training~\cite{ganin2015unsupervised, tzeng2017adversarial}, transfer learning~\cite{ganin2015unsupervised, tzeng2017adversarial}, and pseudo-labeling techniques~\cite{tarvainen2017mean, liu2021unbiased, li2022cross, cai2019exploring, deng2021unbiased}.

Although domain adaptation has been extensively explored in fields such as action recognition~\cite{Plizzari2023, munro_multimodal_domain_2019}, person reidentification~\cite{choudhary2021domain}, video retrieval~\cite{Munro2021DomainAI}, ego-exo adaptation~\cite{Li2021EgoExoTV}, and object detection~\cite{pasqualino2021unsupervised}, little attention has been paid to applying DA to hand-object interaction detection task. One of the key challenges in applying domain adaptation to HOI detection is the substantial domain gap between synthetic and real-world data. Models trained on synthetic datasets often face difficulties when deployed in real-world scenarios due to discrepancies between the generated data and real-world data. It is particularly challenging to create synthetic interactions that accurately reflect the complexity and variability of real-world hand-object interactions. Factors such as natural hand poses, object manipulation, and the diversity of real environments, ranging from kitchens to industrial settings, are difficult to replicate in synthetic datasets. These differences make it difficult for models to generalize effectively, highlighting the need for domain adaptation techniques to bridge the gap and improve performance in real-world applications.

To address this gap, we introduce \textit{HOI-Synth}, a novel benchmark comprising real images, paired with photorealistic labeled synthetic images automatically annotated with labels useful for hand-object detection, such as 2D bounding boxes, semantic segmentation masks, depth maps, and hand-object relations. \textit{HOI-Synth} is specifically designed to facilitate the exploration of domain adaptation techniques for hand-object interaction detection.

\section{The HOI-Synth Benchmark} \label{sec:benchmark}
This section introduces the \textit{HOI-Synth} benchmark, developed to support research on synthetic-to-real domain adaptation for egocentric hand-object interaction detection. \textit{HOI-Synth} combines three egocentric hand-object interaction benchmarks with synthetic data generated through our hand-object interaction data generation pipeline. \textit{HOI-Synth}, along with the data generation tool, is publicly released to encourage further research on using synthetic data for egocentric HOI detection.

\subsection{HOI-Synth Data Generation Pipeline} \label{sec:simulator}
\begin{figure}[t]
     \centering
     \includegraphics[width=\linewidth]{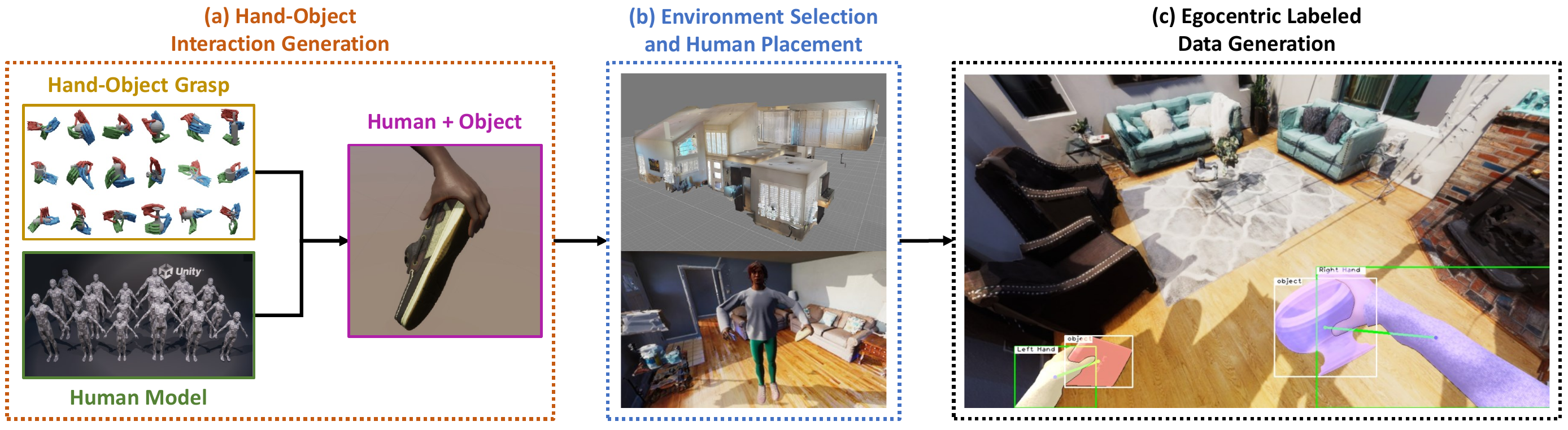}
     \caption{\textbf{The proposed data generation pipeline.} (a) An object-grasp pair is selected from \textit{DexGraspNet}~\cite{wang2023dexgraspnet} and integrated with a randomly generated human model. (b) The human + object model is placed in an environment randomly selected from the \textit{Habitat-Matterport 3D} dataset~\cite{ramakrishnan2021habitat}. (c) Egocentric data of hand-object interactions is generated and automatically labeled. Labels include bounding boxes and segmentation masks of hands and interacted objects, contact-state, and hand-object relations.}
     \label{fig:unity_tool}
 \end{figure}

Figure~\ref{fig:unity_tool} shows the proposed data generation pipeline, which comprises three main blocks: \textit{Hand-Object Interaction Generation}, \textit{Environment Selection and Human Placement}, and \textit{Egocentric Labeled Data Generation}. Our pipeline exploits state-of-the-art datasets and different developed modules to generate realistic egocentric hand-object interaction data~\cite{wang2023dexgraspnet,unity_synthetichumans_2022,ramakrishnan2021habitat,shadowhand}. A detailed description of the pipeline follows.

\begin{enumerate}
    \item \textbf{Hand-Object Interaction Generation:} The generation process begins by selecting a random hand-object grasp from the \textit{DexGraspNet} dataset~\cite{wang2023dexgraspnet}, which includes over 1.32 million robotic grasps with 5,355 untextured, convex-decomposed 3D models from popular datasets like \textit{ShapeNet}~\cite{chang2015shapenet, savva2015semantically}, \textit{YCB}~\cite{calli2017yale}, \textit{Big-BIRD}~\cite{singh2014bigbird}, \textit{Grasp}~\cite{kappler2015leveraging}, \textit{KIT}~\cite{kasper2012kit}, and \textit{Google's Scanned Objects}~\cite{downs2022google}. To enhance realism, we retrieved the original textures of the objects from these datasets. The hand pose from \textit{DexGraspNet} is then converted from the \textit{ShadowHand} rigging~\cite{shadowhand} to match the format of \textit{SyntheticHumans}~\cite{unity_synthetichumans_2022}.
    \begin{itemize}
        \item \textbf{Human Model Generation:} Once the grasp is selected, it is applied to a randomly generated human model using \textit{SyntheticHumans}~\cite{unity_synthetichumans_2022}, which provides customizable 3D human models with parameters like ethnicity, clothing, age, height, weight, and gender (Figure~\ref{fig:unity_tool}-a).
    \end{itemize}
    \item \textbf{Environment Selection, Placement, and Camera Model:} The human model is then placed in a randomly selected environment from the \textit{HM3D} dataset~\cite{ramakrishnan2021habitat} (Figure~\ref{fig:unity_tool}-b). To ensure realistic placement and avoid collisions with scene geometry (e.g., walls or furniture), we utilize a navigation mesh (NavMesh~\footnote{\url{https://docs.unity3d.com/6000.3/Documentation/ScriptReference/AI.NavMesh.html}}) to constrain the human model's position to valid walkable areas. 
    \newline\textit{Camera Parameters:} A virtual camera is attached to the head bone of the human model to simulate a first-person perspective. The camera is positioned at eye level with a relative offset ($y=+0.1m, z=+0.2m$) to align with the virtual eyes. To replicate the wide-angle of standard wearable cameras used, we set a fixed FOV of $94.4$. Furthermore, to introduce natural variability in head orientation and avoid repetitive framing, a randomization module modifies the camera's rotation for each generated scene.
    \item \textbf{Egocentric Labeled Data Generation and Rendering:} After placing the human-object model in the environment, the data generation is performed using the \textit{High Definition Render Pipeline} (HDRP) within Unity, which ensures high-fidelity visual rendering. To introduce realistic variations, we implemented a custom lighting randomizer that dynamically adjusts light intensity, color, and direction.
    Annotations are generated automatically using a customized version of the \textit{Unity Perception}~\footnote{https://docs.unity3d.com/Packages/com.unity.perception@1.0/manual/index.html} package. This enables pixel-perfect ground truth generation, handling occlusions by rendering semantic segmentation masks only for visible pixels.
    The simulator renders static RGB frames at a resolution of $1280 \times 720$ pixels, alongside 2D bounding boxes, segmentation masks for hands and objects, contact states, and hand-object interactions (Figure~\ref{fig:unity_tool}-c). The pipeline supports various augmentations, including changes in lighting conditions and textures, as well as effects like motion blur or noise.
    Furthermore, the simulator offers additional capabilities, including depth maps, 3D bounding boxes, and 3D pose generation, as well as the integration of multiple cameras from different viewpoints. Although this work focused on egocentric RGB frames, future research could leverage the simulator's potential by exploring diverse viewpoints, multimodal data, and 3D annotations.
    Finally, the simulator allows the generation of data by selecting specific aspects of the scene, enabling users to focus on particular objects, hand poses, and environments tailored to their research needs. The interaction probability for each hand can be configured independently, providing full control over the handedness distribution (i.e., left, right, and bi-manual) of the generated data. This targeted approach ensures that the synthetic data aligns closely with specific scenarios, facilitating a more fine-grained analysis of hand-object interactions. By customizing these elements, researchers can generate diverse datasets that reflect various real-world conditions, enhancing the applicability of the models developed using this data (as discussed in Section~\ref{sec:align_synth}).
    \begin{itemize}
        \item \textbf{Scene Composition and Clutter.} To further increase visual complexity, the pipeline includes a clutter generation module that adds random \textit{distractor objects} from the scene object pool. These objects are not part of the interaction and mainly contribute to background clutter and occlusions.
    \end{itemize}
\end{enumerate}

\paragraph{Frame-level generation and temporal coherence}
In this work, our data generation pipeline is used to generate independent egocentric frames. Following the standard HOS formulation~\cite{VISOR2022}, both training and evaluation are performed at the single-frame level, using COCO mask AP metrics, and models do not exploit temporal cues from real videos. Consequently, our synthetic data are sampled frame-wise, and no temporal coherence (e.g., consistent hand–object motion or camera trajectories) is enforced between consecutive images.

\begin{figure}[!ht]
     \centering
     \includegraphics[width=\linewidth]{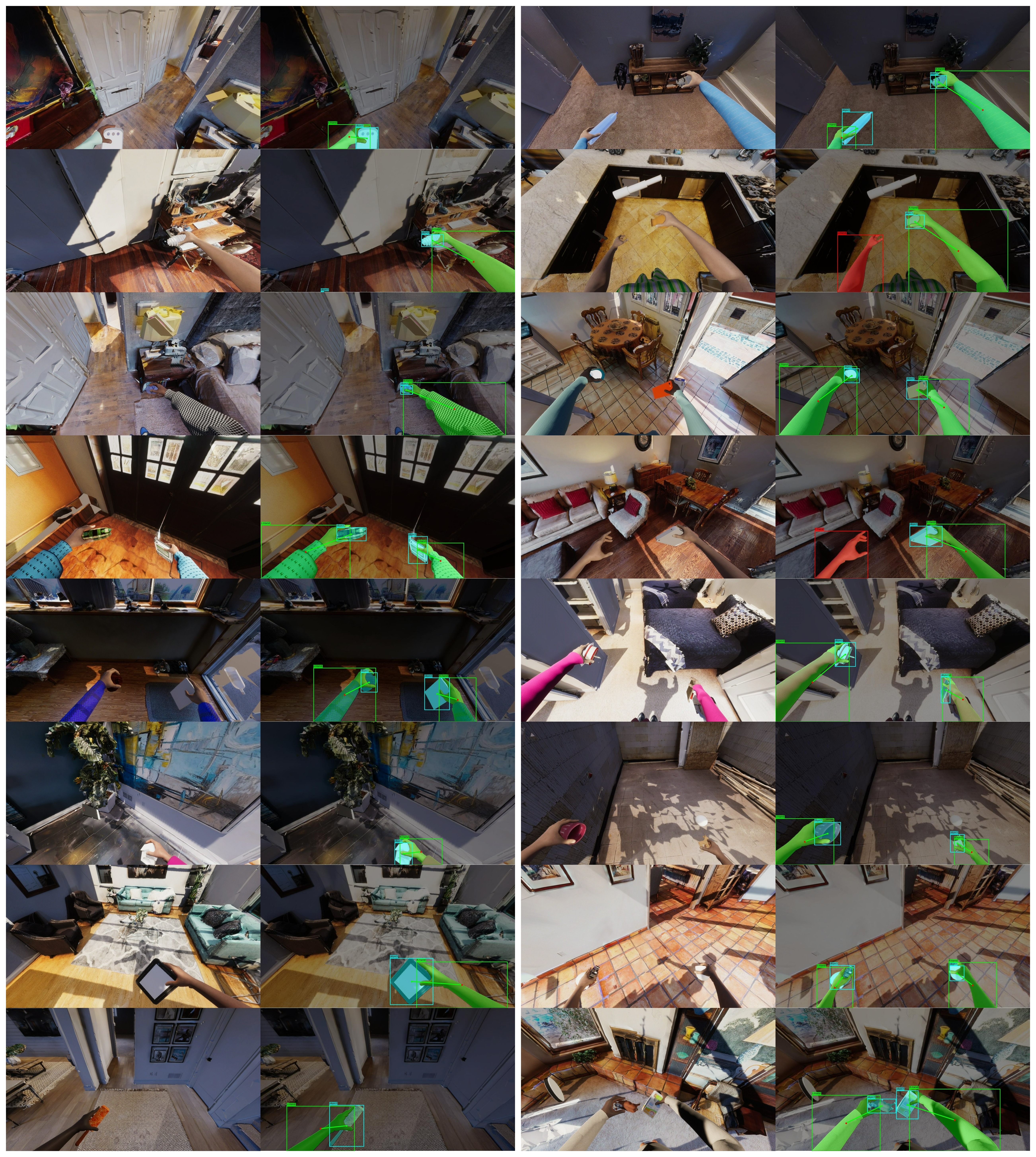}
     \caption{Qualitative examples generated by our \textit{HOI-Synth Data Generation Pipeline}. For each sample, the pipeline automatically generates rich ground-truth annotations, including 2D bounding boxes, pixel-wise segmentation masks, and interaction metadata describing the relationship between hands and manipulated objects.}
     \label{fig:examples_sim}
 \end{figure}

\paragraph{Visual Analysis of the Generated Data}
Figure~\ref{fig:examples_sim} shows qualitative examples of the synthetic samples generated by our pipeline. By varying human models, objects, and environments, the generated data covers a wide range of scene configurations. The rendered frames exhibit diverse visual conditions, such as changes in illumination, background clutter, and self-occlusions, which introduce challenges commonly encountered in egocentric scenarios. The Figure further shows that the automatically generated annotations remain spatially consistent at the pixel level across these variations, even in the presence of complex hand–object interactions. Importantly, while the generated images are not intended to achieve indistinguishable photorealism, our results indicate that capturing sufficient visual variability and structural complexity is more critical for effective sim-to-real transfer than maximizing visual realism alone.

\subsection{Datasets}\label{sec:datasets}
By integrating automatically labeled synthetic data produced through the proposed generation pipeline, the \textit{HOI-Synth} benchmark extends three datasets for egocentric hand-object interaction detection: \textit{VISOR}~\cite{VISOR2022}, \textit{EgoHOS}~\cite{EgoHos_jianbo_eccv22}, and \textit{ENIGMA-51}~\cite{ragusa2024enigma}, described in the following.

\begin{figure*}[t]
    \centering
    \includegraphics[width=\linewidth]{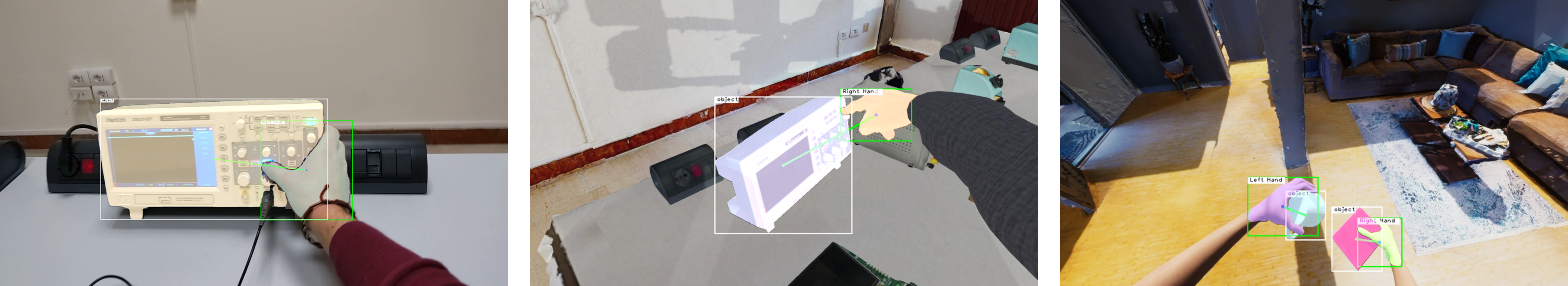}
    \caption{A ENIGMA-51 image (left), a synthetic in-domain image (center), and a synthetic out-domain image (right).}
    \label{fig:in-domain}
\end{figure*}

\begin{itemize}
    \item \textbf{\textit{VISOR}~\cite{VISOR2022}} includes 36 hours of egocentric videos from \textit{EPIC-KITCHENS-100}~\cite{Damen2021RESCALING}, with 32,857 training images and pixel-wise annotations for 42,787 hand-object interactions. We augment this dataset with 30,259 synthetic images, covering 45,353 additional HOIs. Additionally, we generated multiple sets of data specifically designed to target individual aspects of hand-object interactions (i.e., objects, grasps, and environments) to better align with the real-world characteristics of \textit{VISOR} data. These aligned datasets enable a more focused analysis of how each HOI aspect contributes to overall model performance in real-world applications. Additional information on the generation process can be found in Section~\ref{sec:align_synth}.
    \item \textbf{\textit{EgoHOS}~\cite{EgoHos_jianbo_eccv22}} contains 8,107 egocentric training images of HOIs, sparsely sampled from videos of \textit{EGO4D}~\cite{Grauman2021Ego4DAT}, \textit{THU-READ}~\cite{Tang2017ActionRI}, \textit{EPIC-KITCHENS}~\cite{damen2018scaling}, and other egocentric videos of escape room activities. The dataset provides pixel-wise annotations for 13,659 hand-object relations. We augment this dataset with 8,107 synthetic images, including 12,129 additional HOIs.
    \item \textbf{\textit{ENIGMA-51}~\cite{ragusa2024enigma}} is an egocentric dataset where participants follow instructions to repair electrical boards in an industrial setting. It consists of 51 videos, approximately 22 hours of footage, and 45,505 labeled images. We used a subset of \textit{ENIGMA-51} labeled for the egocentric hand-object interaction detection task. This subset includes 3,479 training images with pixel-level annotations for 13,659 hand-object interactions, along with 3D models of the objects and the laboratory environment. We extend this dataset with two sets of synthetic images: one \textit{in-domain} and one \textit{out-domain}. The in-domain set (Figure~\ref{fig:in-domain}-center) is generated using the provided 3D models of the objects and environment, ensuring alignment with the real-world data. The out-domain set contains synthetic images of hand-object interactions in generic environments with generic objects, similar to those used for \textit{VISOR} and \textit{EgoHOS} (Figure~\ref{fig:in-domain}-right). 
\end{itemize}

Table~\ref{tab:datasets} shows the statistics for the training data of the \textit{HOI-Synth} benchmark, detailing the number of real and synthetic images, annotated hands, objects, and hand-object interactions. The official validation and test sets of each dataset are used for evaluation purposes.

\subsubsection{Object Overlap Analysis}
An important step in aligning synthetic data with real-world datasets is maximizing the overlap between object classes covered in real and synthetic data. The overlap percentages between synthetic data in \textit{HOI-Synth} and the real datasets are as follows: $46.6\%$ for \textit{VISOR}, $29.1\%$ for \textit{EgoHOS}, and $0.4\%$ for \textit{ENIGMA-51}. These percentages reflect the extent to which object classes in the synthetic data are present in each real dataset. The results shown in Sections~\ref{sec:align_synth} and \ref{sec:enigma_results} highlight that the effectiveness of synthetic data correlates strongly with the degree of overlap in object classes, underscoring the importance of tailoring synthetic data for specific domains, especially for datasets where the overlap is minimal.

\subsubsection{Data Scale Analysis}
The \textit{ENIGMA-51} dataset required approximately $6\times$ more synthetic images than real ones (20,321 out-domain vs. 3,479 real images) due to the small size of the real training set and its industrial nature, which exhibits minimal object overlap ($0.4\%$) with out-domain synthetic data. Our analysis shows that such large synthetic volumes ($\sim$20k images) are necessary for performance gains, particularly when real data coverage is limited and domain overlap is minimal.

\begin{table*}[t]
    \centering
    \caption{Statistics of the training sets included in our \textit{HOI-Synth} benchmark.}\label{tab:datasets}
    \resizebox{\linewidth}{!}{
        \begin{tabular}{lrrrr}
        \hline
        \textbf{Dataset}                    & \textbf{Images} & \textbf{Hands} & \textbf{Objects} & \textbf{HOI} \\
        \hline               
        VISOR~\cite{VISOR2022}              & 32,857          & 52,906         & 42,785           & 42,787       \\
        Synthetic                           & 30,259          & 60,098         & 45,219           & 45,353       \\ 
        Synthetic aligned Objects           & 4,951           & 9,810          & 8,311            & 8,263        \\
        Synthetic aligned Grasps            & 5,000           & 9,899          & 8,355            & 8,306        \\
        Synthetic aligned Environments      & 4,938           & 9,789          & 8,034            & 8,002       \\
        Synthetic aligned All               & 31,000          & 61,408         & 52,279           & 52,045      \\
        \hline 
        EgoHOS~\cite{EgoHos_jianbo_eccv22} & 8,107           & 15,015         & 11,393           & 13,659       \\ 
        Synthetic                           & 8,107           & 16,101         & 12,170           & 12,129       \\ 
        \hline
        ENIGMA-51~\cite{ragusa2024enigma} & 3,479           & 5,075          & 4,343            & 4,344        \\ 
        Synthetic-In-Domain                 & 16,773          & 25,444         & 16,637           & 16,773       \\     
        Synthetic-Out-domain             & 20,321          & 40,135         & 27,499           & 27,370       \\ 
        \hline
        \end{tabular}
    }
\end{table*}

\section{Experimental Analysis and Results} \label{sec:results}
\begin{figure*}[!ht]
    \centering
    \includegraphics[width=\linewidth]{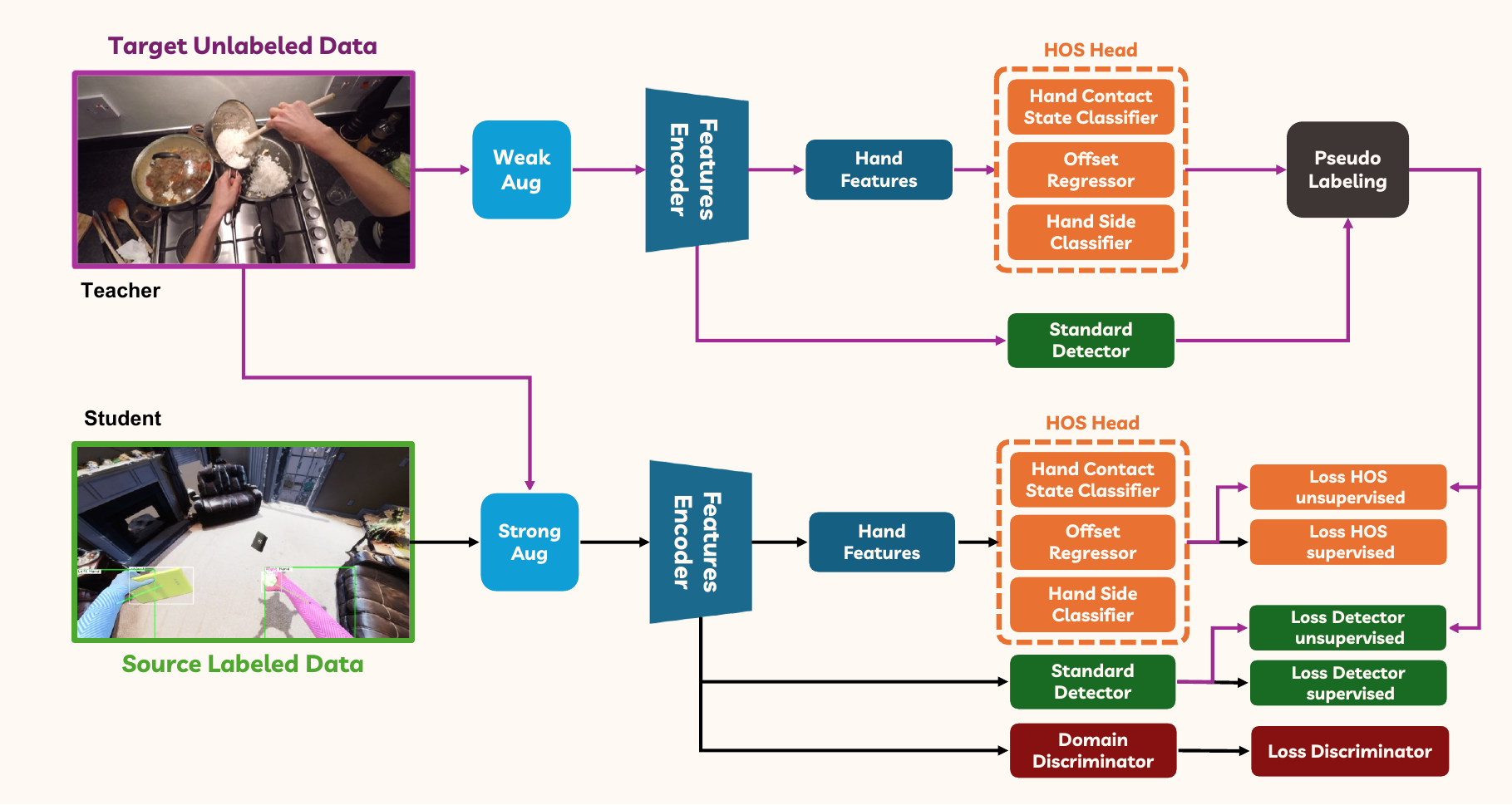}
    \caption{The architecture of the domain adaptation approach used in our analysis. The method is based on the \textit{Adaptive Teacher}~\cite{li2022cross} framework, extended with HOS recognition modules~\cite{VISOR2022}.}
    \label{fig:hos_at_arch}
\end{figure*}

We adopt \textit{VISOR HOS}~\cite{VISOR2022} as the baseline for our experiments. This method extends the \textit{PointRend}~\cite{kirillov2020pointrend} instance segmentation network by introducing three new modules, implemented as standard MLPs: one for identifying the hand's side, another for determining the contact state  (``contact'' or ``no contact''), and a third for estimating an offset vector to link the hand with the interacted object. We explore five instances of the hand-object interaction detector based on the \textit{VISOR HOS} approach:

\begin{itemize}
    \item \textbf{Synthetic-Only} This variant of the \textit{VISOR HOS} model is trained exclusively on synthetic data and evaluated directly on real data. This approach aims to determine whether synthetic data can fully substitute for real data.
    \item \textbf{Unsupervised Domain Adaptation (UDA)} This approach combines the \textit{VISOR HOS} architecture and the \textit{Adaptive Teacher}~\cite{li2022cross} (AT), an unsupervised domain adaptation framework for cross-domain object detection. We modified AT for HOI detection by including the additional modules for estimating the hand side, contact state, offset vector, and segmentation masks. The model is trained using labeled synthetic and unlabeled real data in an unsupervised domain adaptation setting and works as follows (see Figure~\ref{fig:hos_at_arch}): 
    \begin{itemize}
        \item \textbf{Teacher Model:} Initially, a real-world unlabeled image is passed to the teacher model. A weak augmentation is applied, introducing soft transformations like slight color adjustments or noise injection, without altering the spatial structure of the image. The teacher model then generates HOI pseudo-labels, which serve as pseudo-ground truth for the student model.
        \item \textbf{Student Model:} The student model receives two inputs: the same unlabeled real-world image and a labeled synthetic image. A strong augmentation is applied to the two images, which include more aggressive transformations, such as significant color shifts, noise addition, and brightness or contrast changes. Note that, no spatial augmentations like cropping, rotation, or flipping are used to maintain the integrity of the pseudo-labels generated by the teacher. The student then uses the pseudo-GT during training for the unlabeled real-world images.
        \item \textbf{Domain Discriminator:} A domain discriminator is incorporated into the pipeline with a \textit{Gradient Reversal Layer}~\cite{ganin2015unsupervised} (GRL). This helps to align the feature distributions of real and synthetic images, reducing the domain gap. The standard binary cross entropy loss is used. 
        \item \textbf{Teacher-Student Update via EMA:} The student model updates the teacher’s weights using a standard \textit{Exponential Moving Average} (EMA). This technique ensures that the teacher’s weights evolve gradually and in parallel with the student model.
    \end{itemize}
     The goal of this approach is to evaluate whether current domain adaptation techniques when paired with high-quality labeled synthetic data, can eliminate the need for labeling real data.
    \item \textbf{Real-Only} This method involves training the \textit{VISOR HOS} model exclusively on labeled real data. We conduct experiments with varying quantities (i.e., 10\%, 25\%, 50\%, and 100\%) of labeled real data to evaluate the extent to which performance is influenced by the scale of training data in the absence of synthetic images. This approach serves as a baseline and represents the typical fully supervised setup.
    \item \textbf{Synthetic + Real} This approach involves pre-training the \textit{VISOR HOS} model on labeled synthetic data followed by fine-tuning on labeled real data. We also vary the amounts of labeled real data (i.e., 10\%, 25\%, 50\%, and 100\%) in our experiments. The goal is to evaluate the effectiveness of labeled synthetic data in reducing the reliance on real data for training, without implementing any specific synth-to-real adaptation.
    \item \textbf{Semi-Supervised Domain Adaptation (SSDA)} This method involves training the AT model using labeled synthetic data, unlabeled real data, and a subset of labeled real data. We test various amounts of labeled real data (i.e., 10\%, 25\%, 50\%). To enable the Adaptive Teacher model to work in a semi-supervised manner, we combine the labeled synthetic data with the labeled real data. The goal of these experiments is to evaluate whether synthetic data can enhance performance when only a subset of the real data is labeled.
    \item \textbf{Fully-Supervised Domain Adaptation (FSDA)} In this method, we train the AT model by combining labeled synthetic data with all available labeled real data while implementing domain adaptation techniques. The goal of this method is to evaluate whether synthetic data can enhance state-of-the-art performance, even when a substantial amount of labeled real data is present.
\end{itemize}
\subsection{Implementation Details}
\subsubsection{Evaluation Measures} Following~\cite{VISOR2022}, we evaluate performance using \textit{COCO Mask AP}~\cite{coco_dataset}. Specifically, we focus on the \textit{Hand + Object} (Overall) AP, which assesses the accuracy of the combined masks of hands and in-contact objects. We also provide a detailed performance breakdown in terms of AP that evaluates specific prediction aspects: \textit{Hand} (H), \textit{Hand + Side} (H+S), \textit{Hand + Contact} (H+C), and \textit{Object} (O). Finally, to further analyze model performance on specific downstream tasks, we compute \textit{F1 scores} for hand contact and side predictions independently of hand segmentation accuracy. This approach allows us to assess the model's ability to predict these aspects of hand-object interactions without influence from segmentation errors.
\subsubsection{Computational Cost and Efficiency}
To ensure reproducibility and provide a clear assessment of resource requirements, we report the computational budget for both synthetic data generation and model training. \\
\noindent\textbf{Data Generation Efficiency} \\
Our simulation pipeline is highly efficient and can be run on readily available hardware. On a laptop equipped with an \textit{NVIDIA RTX 2080} GPU and \textit{32 GB} of RAM, the simulator generates fully annotated synthetic samples at a rate of approximately $10,000$ images every \textasciitilde$1.5$ hours. This efficiency allows the creation of large-scale, diverse datasets with minimal time investment compared to manual collection and annotation of real-world data. \\
\noindent\textbf{Training Cost} \\
Model training was performed on a server node with \textbf{4 NVIDIA A30 GPUs (24~GB VRAM each)}. Training the full framework is memory-intensive due to the teacher-student architecture, which processes both labeled and unlabeled images simultaneously. For representative experiments, each batch contains 4 labeled and 4 unlabeled images per GPU, fully utilizing the 24~GB VRAM. A full training run (e.g., FSDA with 80k iterations) requires approximately 40~hours on this multi-GPU setup. Additional details on batch sizes, iterations, and per-experiment configurations are provided in our publicly released repository: \url{https://github.com/fpv-iplab/HOI-Synth/tree/baseline-code}.\\
\noindent\textbf{Inference Efficiency} \\
During inference, only the student model is deployed, ensuring efficient evaluation. Each image is processed in approximately \textbf{0.08~s} (\textasciitilde 12~FPS) while maintaining low GPU memory usage.

\begin{table*}[!t]
	\centering
        \caption{Results on the \textit{VISOR} validation set considering different real data settings available in training. Gray rows indicate {\sethlcolor{gray!30}\hl{baseline models}} in each configuration, while white rows highlight models trained with synthetic and real data. In each group, the \textbf{best results} are in bold, while the \underline{best results among the models trained with synthetic and real data} are underlined. \textcolor{blue}{Overall enhancements} are shown in blue, indicating improvements of the models trained with synthetic and real data over the {\sethlcolor{gray!30}\hl{baseline}.}}
		
	\begin{minipage}{\linewidth} 
		\centering
		\textbf{a) Unsupervised Setting} \\
		\resizebox{\linewidth}{!}{
			    
			\begin{tabular}{c|c|c|cccc}
				\hline
				\textbf{\% Real Labeled Data} & \textbf{Approach}              & \textbf{Overall}                          & H                                         & H+S                                       & H+C                                       & O                                        \\
				\hline
				\multirow{2}{*}{0\%}   & \cellcolor{gray!30} Synthetic-Only     & \cellcolor{gray!30} 09.88 
				& \cellcolor{gray!30} 28.41                           & \cellcolor{gray!30} 24.89                           & \cellcolor{gray!30} 08.64                           & \cellcolor{gray!30} 01.23                                                     \\ 
				                              &  UDA &  \textbf{33.33} &  \textbf{80.16} &  \textbf{65.98} &  \textbf{33.47} &  \textbf{8.35} \\ 
				\hline
				\multicolumn{2}{c|}{Absolute Improvement}   & $\textcolor{blue}{\textbf{+23.45}}$& $+51.75$ & $+41.09$ & $+24.83$ & $+7.12$  \\

				\hline
							      
			\end{tabular} 
		}
	\end{minipage}

	\vspace{1em}
	        
	\begin{minipage}{\linewidth} 
		\centering
		\textbf{b) Semi-supervised Setting} \\
		\resizebox{\linewidth}{!}{
			       
			\begin{tabular}{c|c|c|cccc}
				\hline
				\textbf{\% Real Labeled Data} & \textbf{Approach}                         & \textbf{Overall}                                      & H                                                     & H+S                                          & H+C                                                   & O                                                     \\
				\hline
				\multirow{3}{*}{\makecell{10\% \\ (3,286 images)}}  & \cellcolor{gray!30} Real-Only          
				& \cellcolor{gray!30} 38.55 & \cellcolor{gray!30} 87.45                           & \cellcolor{gray!30} \textbf{83.27}                  & \cellcolor{gray!30} 51.98                           & \cellcolor{gray!30} 19.47                                                     \\ 
				                              &  Synthetic+Real &  37.62                      &  86.39                      &  \underline{82.85} &  \textbf{\underline{52.25}} &  \textbf{\underline{23.03}} \\ 
				                              &  SSDA           &  \textbf{\underline{44.22}} &  \textbf{\underline{89.05}} &   80.77            &  46.83                      &  20.41                      \\ 
				\hline
				\multicolumn{2}{c|}{Absolute Improvement}  & $\textcolor{blue}{\textbf{+5.67}}$ & $+1.60$ & $-0.42$ & $+0.27$ & $+3.56$  \\
				
							
				\hline
				\multirow{3}{*}{\makecell{25\% \\ (8,215 images)}}  & \cellcolor{gray!30} Real-Only           
				& \cellcolor{gray!30} 37.90
				& \cellcolor{gray!30} 90.14                           & \cellcolor{gray!30} \textbf{85.66}                  & \cellcolor{gray!30} 53.99                           & \cellcolor{gray!30} 17.85                                                     \\ 
				&  Synthetic+Real 
				&  38.19 
				&  89.98                      &  \underline{84.67}          &  \textbf{\underline{55.88}} &  18.49                                          \\ 
				&  SSDA   &  \textbf{\underline{45.55}}        
				&  \textbf{\underline{90.37}} &  84.42                      &  52.59                      &  \textbf{\underline{22.15}}  \\ 
				\hline
				\multicolumn{2}{c|}{Absolute Improvement} & $\textcolor{blue}{\textbf{+7.65}}$  & $+0.23$ & $-0.99$ & $+1.89$ & $+4.30$ \\
				\hline
				\multirow{3}{*}{\makecell{50\% \\ (16,429 images)}}  & \cellcolor{gray!30} Real-Only   
				& \cellcolor{gray!30} 38.15
				& \cellcolor{gray!30} 91.16                           & \cellcolor{gray!30} \textbf{86.05}                  & \cellcolor{gray!30} 52.28                           & \cellcolor{gray!30} 17.92                                                     \\ 
				&  Synthetic+Real 
				&  43.52 
				&  \textbf{\underline{91.34}} &  \underline{85.85}          &  54.09                      &  19.06                                           \\
				&  SSDA          
				&  \textbf{\underline{46.47}}
				&  90.94                      &  85.73                      &  \textbf{\underline{58.02}} &  \textbf{\underline{23.49}} \\ 
				\hline
				\multicolumn{2}{c|}{Absolute Improvement} & $\textcolor{blue}{\textbf{+8.32}}$  & $+0.18$ & $-0.20$ & $+5.74$ & $+5.57$  \\
				   
				\hline  
			\end{tabular}
		}
	\end{minipage}

	\vspace{1em}
	       
	\begin{minipage}{\linewidth} 
		\centering
		\textbf{c) Fully-supervised Setting} \\
		\resizebox{\linewidth}{!}{
			       
			\begin{tabular}{c|c|c|cccc}
				\hline
				\textbf{\% Real Labeled Data} & \textbf{Approach} & \textbf{Overall} & H & H+S & H+C & O \\
				\hline
				\multirow{3}{*}{\makecell{100\% \\ (32,857 images)}}
				& \cellcolor{gray!30} Real-Only 
				& \cellcolor{gray!30} 45.33& \cellcolor{gray!30} \textbf{92.25}                  & \cellcolor{gray!30} 88.54                           & \cellcolor{gray!30} \textbf{59.24}                  & \cellcolor{gray!30} 24.23                                                      \\ 
				&  Synthetic+Real 
				&  44.52&  91.45                      &  \textbf{\underline{88.94}} &  56.55                      &  \textbf{\underline{27.77}}                       \\
				&  FSDA   &  \textbf{\underline{46.48}}       
				&  \underline{91.83}          &  87.65                      &  \underline{57.63}          &  24.03                       \\
				\hline
				\multicolumn{2}{c|}{Absolute Improvement} & $\textcolor{blue}{\textbf{+1.15}}$ & $-0.42$  & $+0.40$ & $-1.61$ & $+3.54$  \\
				
				\hline
				    			      
			\end{tabular}
		}
	\end{minipage}	  
	\label{tab:hos_visor}
 
\end{table*}

\subsection{Results on VISOR} \label{sec:visor_results}
The results obtained on the validation\footnote{We used the validation set since the annotations for the test set are not publicly available.} set of \textit{VISOR}~\cite{VISOR2022} are shown in Table~\ref{tab:hos_visor}. It is important to note that, in our implementation, the results of the HOS model differ from those reported in~\cite{VISOR2022} due to our use of a batch size of 4, which represents the largest batch size achievable with domain adaptation models in our configuration, ensuring fair comparisons. 

Training the baseline model exclusively on synthetic data (Table~\ref{tab:hos_visor}-a) results in poor performance, with an \textit{Overall} AP of 9.88\%. This is significantly lower than the 45.33\% achieved using the full set of real training data (Table~\ref{tab:hos_visor}-c). However, using UDA techniques results in significant enhancements in model performance across all evaluation metrics, with an increase of +23.45\% in \textit{Overall} AP. Hand-dependent APs improved by +51.75\%, +41.09\%, and +24.83\%, while \textit{Object} AP increased by +7.12\%. These results confirm the effectiveness of synthetic data when combined with UDA in bridging the synthetic-to-real domain gap.

In the semi-supervised setting (Table~\ref{tab:hos_visor}-b), models trained on both synthetic and real data consistently outperform baselines trained exclusively on real data, regardless of the percentage of labeled real training data (10\%, 25\%, and 50\%). Specifically, the \textit{SSDA} approach achieves improvements of +5.67\%, +7.65\%, and +8.32\% in \textit{Overall} AP, respectively. Furthermore, \textit{Object} AP (O) increases by +3.59\%, +4.30\%, and +5.57\%, and \textit{H+C} AP improves by +0.27\%, +1.89\%, and +5.74\%, demonstrating the effectiveness of synthetic data in enhancing object detection and hand contact state prediction. Regarding \textit{H} and \textit{H+S} metrics, improvements are relatively minor, ranging from [-0.99, +1.6]. This is likely due to the limited potential for improvement in these metrics (already in the 80\%-90\% range) and real-only models' tendency to overfit these sub-tasks, leading to suboptimal overall performance.

In a fully-supervised configuration (Table~\ref{tab:hos_visor}-c), the \textit{FSDA} method improves upon the \textit{Real-Only} approach by +1.15\% in \textit{Overall} AP and +3.54\% in \textit{Object} AP, achieving comparable performance on \textit{H}, \textit{H+S}, and \textit{H+C} metrics, with improvements ranging from [-1.61, +0.4]. Notably, with just 25\% of labeled real training data (8,215 images), the \textit{SSDA} approach achieves an \textit{Overall} AP of 45.55\%, i.e., a +7.65\% gain over \textit{Real-Only} and even a +0.22\% improvement compared to the fully supervised baseline trained with 100\% of all labeled real data (32,857 images).

\begin{figure*}[!ht]
    \centering
    \includegraphics[width=\linewidth]{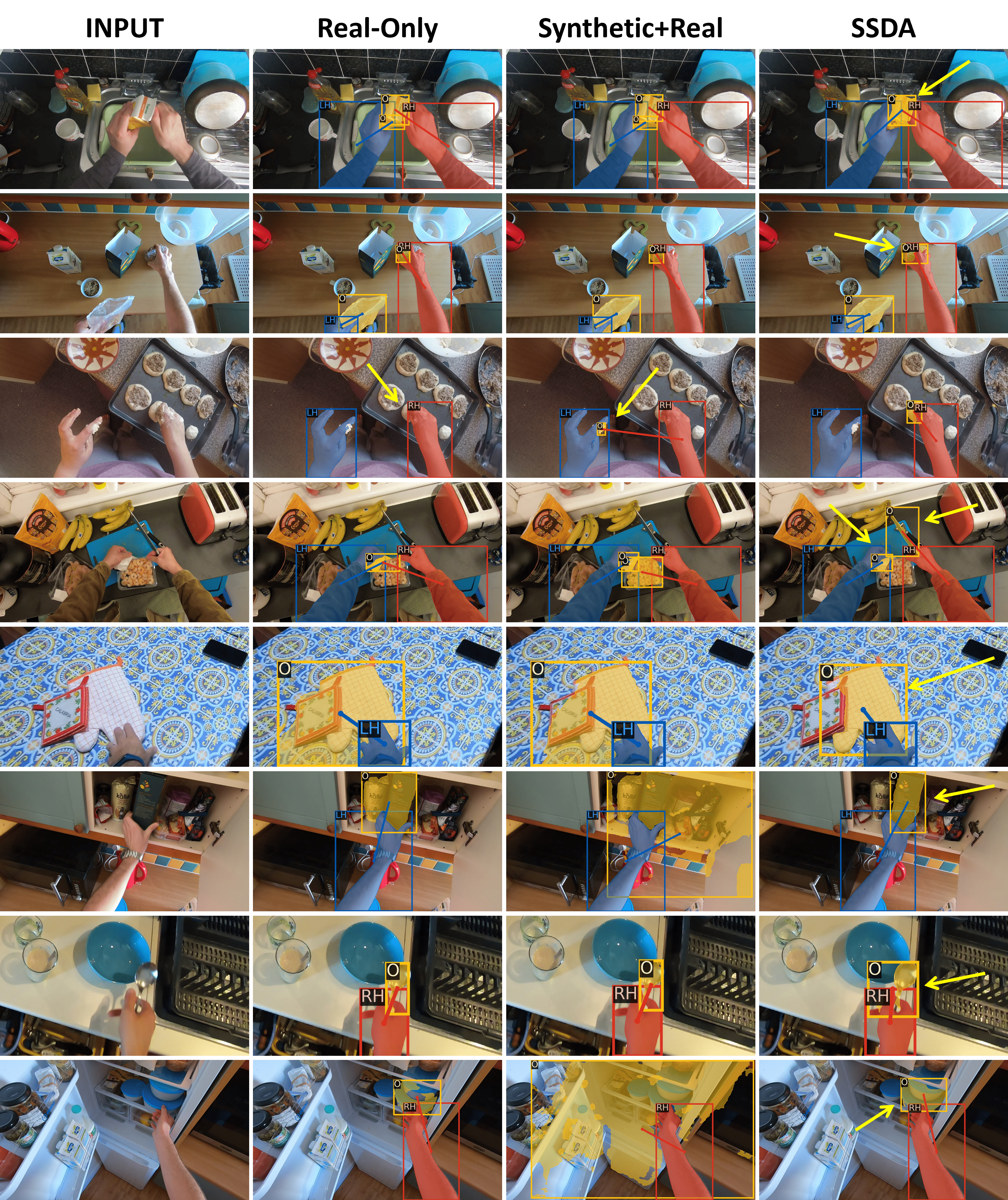}
    \caption{Qualitative examples on \textit{VISOR}. Columns show predictions from \textit{Real-Only}, \textit{Synthetic+Real}, and \textit{SSDA} models. Colors denote the left hand (blue), right hand (red), and interacted objects (yellow). Yellow arrows highlight regions of interest where the SSDA model demonstrates superior performance compared to the other baselines.}
    \label{fig:qualitative_examples_visor}
\end{figure*}

Figure~\ref{fig:qualitative_examples_visor} presents qualitative comparisons on the \textit{VISOR} validation set, with yellow arrows highlighting specific regions of improvement across different scenarios. \ \textbf{1) Segmentation Consistency (Rows 1-2):} The \textit{Real-Only} (second column) and \textit{Synthetic+Real} (third column) baselines often produce fragmented masks or incomplete object segmentations. In contrast, the \textit{SSDA} model (fourth column) generates more accurate masks. \ \textbf{2) Small and Thin Objects (Rows 3, 4):} The baselines struggle significantly with small or thin items, whereas \textit{SSDA} correctly recovers both. \ \textbf{3) Distractor Robustness (Row 5):} In scenes with visually similar background objects (e.g., pot holders next to an oven mitt), baselines are easily confused. The \textit{SSDA} model demonstrates superior robustness, correctly isolating the active hand-object interaction from background distractors. \ \textbf{4) Low-Light and Clutter (Rows 6, 8):} The last examples show challenging conditions such as dark cabinet interiors and cluttered fridge shelves. While the \textit{Real-Only} model often fails to detect objects in these low-visibility areas, \textit{SSDA} correctly recognizes objects even in these low-visibility conditions.


\clearpage

\begin{table*}[t]
    \centering
    \caption{Results of different semi-supervised adaptation approaches trained with synthetic data and $25\%$ \textit{VISOR} labeled training data.}
    \resizebox{0.7\linewidth}{!}{
        \begin{tabular}{l|c|cccc}
            \hline
            \textbf{Method}                     & \textbf{Overall}  & H                 & H+S               & H+C               & O                 \\
            \hline               
            Synthetic + Real                    & 38.19             & 89.98             & \textbf{84.67}    & 55.88             & 18.49             \\
            \hline
            MT~\cite{tarvainen2017mean}         & 43.69             & 88.78             & 84.40             & \textbf{60.94}    & 21.89             \\ 
            MT+GRL~\cite{ganin2015unsupervised} & 43.97             & 88.64             & 84.27             & \underline{58.21} & 21.82             \\ 
            UT~\cite{liu2021unbiased}           & \underline{44.32} & \textbf{90.60}    & \underline{84.49} & 52.55             & \underline{22.11} \\ 
            AT~\cite{li2022cross}               & \textbf{45.55}    & \underline{90.37} & 84.42             & 52.59             & \textbf{22.15}    \\ 
            \hline
        \end{tabular}
    }
    \label{tab:comparison_da_visor}
\end{table*}

\subsubsection{Benchmark of Domain Adaptation Approaches} 
Table~\ref{tab:comparison_da_visor} compares various semi-supervised domain adaptation approaches on \textit{VISOR}, specifically using the setup where only 25\% of real training data is labeled. This configuration is chosen to evaluate the performance of SSDA approaches under limited labeled data, which represents a common scenario in domain adaptation tasks. We compare four approaches: \textit{Mean Teacher~\cite{tarvainen2017mean}} (MT), \textit{Mean Teacher + Adversarial Loss~\cite{ganin2015unsupervised}} (MT+GRL), \textit{Unbiased Teacher~\cite{liu2021unbiased}} (UT), \textit{Adaptive Teacher~\cite{li2022cross}} (AT).

\textit{MT} and \textit{MT+GRL} performed poorly in \textit{Overall AP} (43.69\% and 43.97\%) and \textit{O} (21.89\% and 21.82\%), but excelled in \textit{AP Hand+Contact} (60.94\% and 58.21\%). This suggests that these methods are particularly good at detecting hand-contact relationships, but struggle with overall object detection. In contrast, \textit{UT} achieved the second-best \textit{Overall AP} (44.32\%) and the best \textit{AP Hand} (90.60\%).  However, \textit{AT} outperformed all the other approaches in \textit{Overall AP} (+1.23\% over \textit{UT}) and \textit{AP Object} (22.15\%). These results highlight AT as the superior approach for the task, although the results from the other methods suggest potential areas for further improvement, especially in fine-tuning the performance across different breakdown metrics.

\subsubsection{Scale of Synthesized Data} 
\begin{figure}[t]
    \centering
    \includegraphics[width=0.8\linewidth]{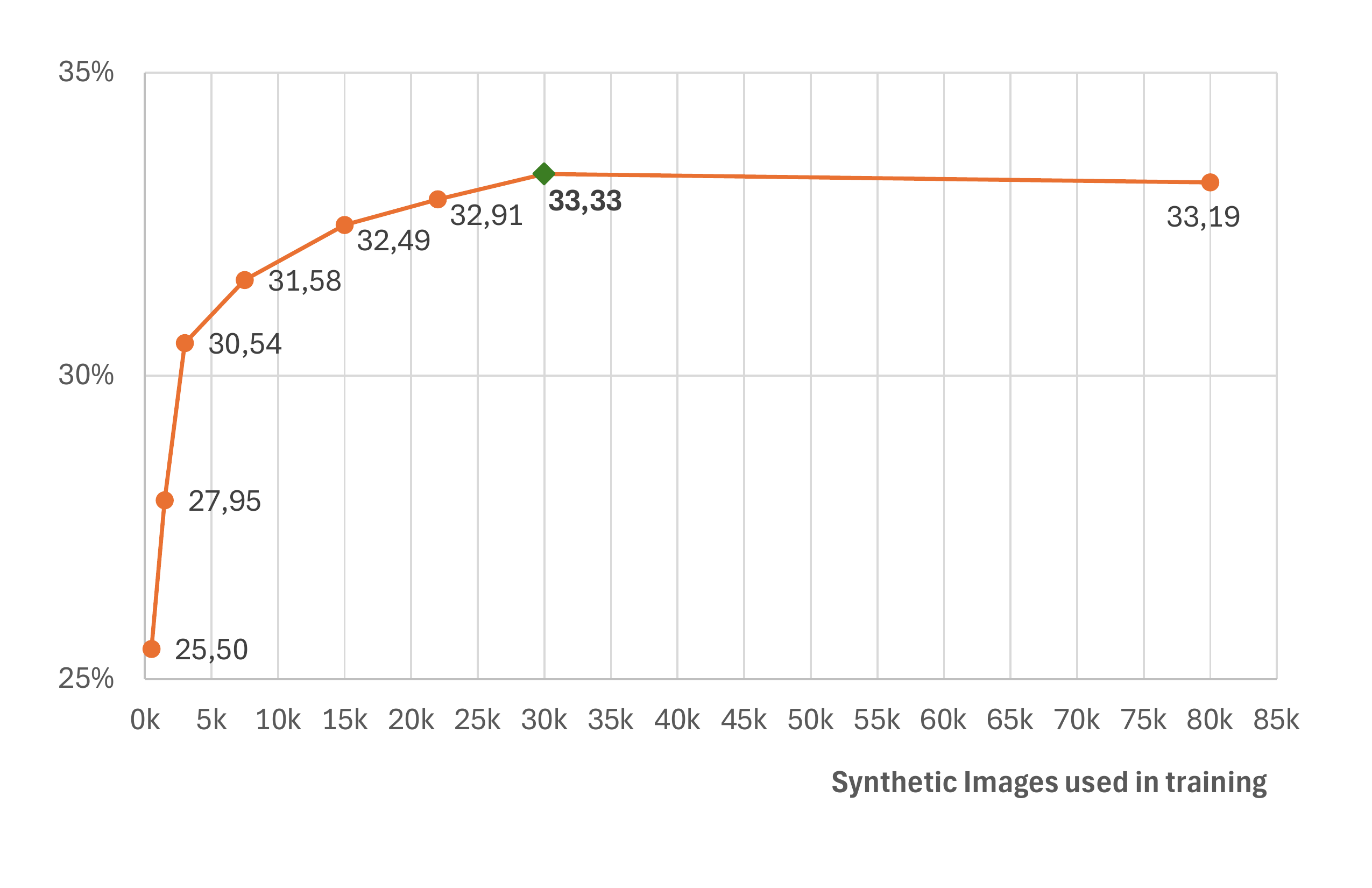}
    \caption{UDA Overall AP on \textit{VISOR} validation set for different amounts of training synthetic data. The green indicator shows the best result.}
    \label{fig:synth_scale}
\end{figure}

Given the ease and cost-effectiveness of generating large quantities of synthetic images with our proposed tool, we aimed to identify the optimal scale of synthetic data needed to maximize or stabilize model performance. To this end, we trained our UDA approach on \textit{VISOR} using various amounts of labeled synthetic data, as shown in Figure~\ref{fig:synth_scale}. The results indicate that the model benefits from incorporating additional synthetic data, reaching a plateau between 22k and 30k training images (33.33\%), and remaining stable with 80,000 synthetic images (33.19\%). This suggests that increasing the data volume beyond this threshold does not lead to significant improvements in performance, highlighting an optimal range for synthetic data that effectively balances efficiency and effectiveness in training.

\subsubsection{Evaluation with Modern Backbone Architectures}
To evaluate the robustness and generality of our findings, we extended our domain adaptation approach beyond the originally employed \textit{ResNet101} backbone by adopting a modern \textit{ConvNeXt}~\cite{liu2022convnet} architecture. Specifically, we adopt the \textit{ConvNeXt-Small} backbone and replicate the same FSDA training settings. To ensure a fair comparison with the baseline, we maintain the same batch size. However, due to the higher computational cost of \textit{ConvNeXt}, we enable mixed-precision training (AMP) to keep memory usage and training time within available hardware constraints. \\
The \textit{ConvNeXt-S} model achieved an Overall AP of $48.36\%$, outperforming the \textit{ResNet101} baseline ($46.48\%$). This result confirms that the effectiveness of our proposed synthetic data strategy is not tied to a specific network configuration but generalizes robustly to modern backbone architectures. To facilitate reproducibility, the full configuration and the trained model used in this experiment are publicly available at \href{https://github.com/fpv-iplab/HOI-Synth/tree/baseline-code?tab=readme-ov-file\#visor--convnext}{our project repository}.

\subsection{Strategies for Generating Synthetic Data Aligned with Real Data}\label{sec:align_synth}
We hypothesize that the effectiveness of synthetic data in enhancing real-world applications is strongly dependent on its alignment with the target real data. While such alignment can be achieved at the scene level for datasets like \textit{ENIGMA-51}, which provide full 3D reconstructions of environments and objects, this assumption does not hold for more heterogeneous benchmarks such as \textit{VISOR} and \textit{EgoHOS}, where such reconstructions are unavailable. To investigate this setting, we explore strategies for improving this alignment at a more granular level, focusing on three key aspects: objects, grasps, and environments. Our experiments, conducted on the \textit{VISOR} dataset, aim to refine the generation of synthetic data to better target the specific characteristics of the real-world data at hand, with the final goal of enhancing the performance of models trained in Unsupervised Domain Adaptation settings. Unlabeled real-world data, along with pre-trained models such as \textit{DINOv2}~\cite{oquab2023dinov2} and \textit{MMPose}~\cite{mmpose2020}, are used to guide the alignment process and refine the synthetic data generation.

\subsubsection{Aligning Object Classes}\label{sec:object_align}
\begin{figure*}[t]
    \centering
    \includegraphics[width=\linewidth]{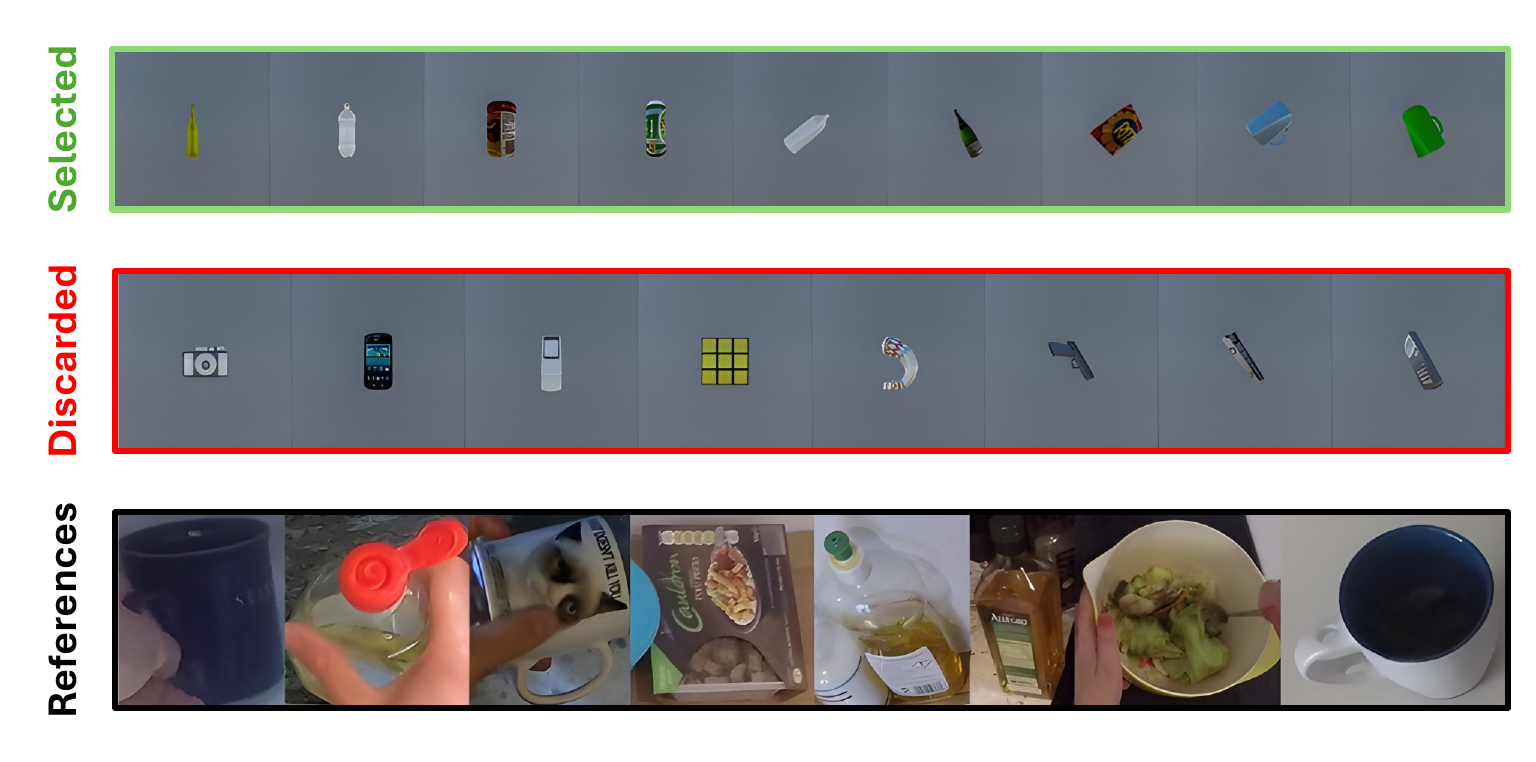}
    \caption{The green row shows the selected synthetic objects aligned with those in \textit{VISOR} (black row). The red row contains the discarded examples, filtered out due to misalignment or poor quality. Note that these objects are common kitchen items, such as mugs, containers, and bottles.}
    \label{fig:objs_aligned}
\end{figure*}

In this experiment, we focused on improving the alignment between synthetic and real objects in the \textit{VISOR} dataset to enhance the performance of the Unsupervised Domain Adaptation (UDA) approach. The alignment process is detailed below. First, we generated a set of images for each synthetic object by placing it in a virtual scene created in our simulator. The virtual camera was rotated around the object to simulate a range of viewpoints, capturing the object's appearance from various angles and orientations. For each image, we then extracted features using \textit{DINOv2}~\cite{oquab2023dinov2}.
Next, we extracted features from the real objects by cropping their regions from the training images using bounding boxes, which were generated using an object detector pre-trained on the \textit{EgoObjects}~\cite{zhu2023egoobjects} dataset. 
After extracting both synthetic and real object features, we applied \textit{DBSCAN}~\cite{ester1996density} to cluster the real object features. Clustering helps in effectively handling outliers, as \textit{DBSCAN} can label anomalous features as noise, preventing them from influencing the alignment process. Additionally, clustering provides a more robust representation of object groups, allowing synthetic features to align not just with individual real objects, but with representative sets of similar real features. Finally, we selected the top 10\% of synthetic objects most similar to the real clusters based on their cosine similarity scores, ensuring that only the most relevant synthetic objects were retained.
Using the selected synthetic objects, we generated a new dataset of hand-object interactions by simulating realistic interactions between hands and objects in our virtual environment. The goal of this process is to generate synthetic data that not only contains a diverse range of objects but also closely matches real-world object distributions, improving the model's ability to perform in real-world applications. See Figure~\ref{fig:objs_aligned} for examples of selected aligned synthetic objects. 

\begin{table*}[t]
	\centering
	\caption{Performance comparison of the UDA Approach with aligned features on the \textit{VISOR} dataset.}
	\resizebox{\linewidth}{!}{
		\begin{tabular}{lcccc|c|ccc}
			\hline
			\textbf{Approach}   & Objects    & Grasps     & Environments & \#Images   & \textbf{Overall} & H              & H+C            & O             \\
			\hline               
			Baseline            &            &            &              & $\sim 5k$  & 31.52            & 79.58          & 30.19          & 7.35          \\
			Target Objects      & \checkmark &            &              & $\sim 5k$  & \textbf{32.60}   & \textbf{82.87} & 34.32          & 7.56          \\
			Target Grasps       &            & \checkmark &              & $\sim 5k$  & 31.84            & 82.01          & 31.95          & \textbf{7.97}         \\
			Target Environments &            &            & \checkmark   & $\sim 5k$  & 31.59            & 81.22          & 32.68          & 7.16          \\
            Target All          & \checkmark & \checkmark & \checkmark   & $\sim 5k$  & 32.59            & 82.86          & \textbf{34.37} & 7.53          \\

			\hline 
			Baseline            &            &            &              & $\sim 30k$ & 33.33            & 80.16          & 33.47          & 8.35          \\
			Target All          & \checkmark & \checkmark & \checkmark   & $\sim 30k$ & \textbf{33.98}             & \textbf{83.20}              & \textbf{34.19}             & \textbf{9.22}       \\
			\hline 
		\end{tabular}
	}
	\label{tab:target_visor}
\end{table*}

The results presented in Table~\ref{tab:target_visor} demonstrate the impact of aligning synthetic and real objects on the performance of our UDA approach. Training the UDA model with aligned data leads to improvements across all evaluation metrics. Specifically, the overall mAP increases from $31.52\%$ to $32.60\%$, indicating that the alignment of synthetic and real objects enhances the model’s general performance. Additionally, we observe improvements in the breakdown metrics: \textit{Hand (H)} increases from $79.58\%$ to $82.87\%$, \textit{Hand+Contact (H+C)} from $30.19\%$ to $34.32\%$, and \textit{Object (O}) from $7.35\%$ to $7.56\%$, further demonstrating the effectiveness of this alignment in improving both hand-object interaction recognition and object detection.

\begin{figure*}[t]
    \centering
    \includegraphics[width=\linewidth]{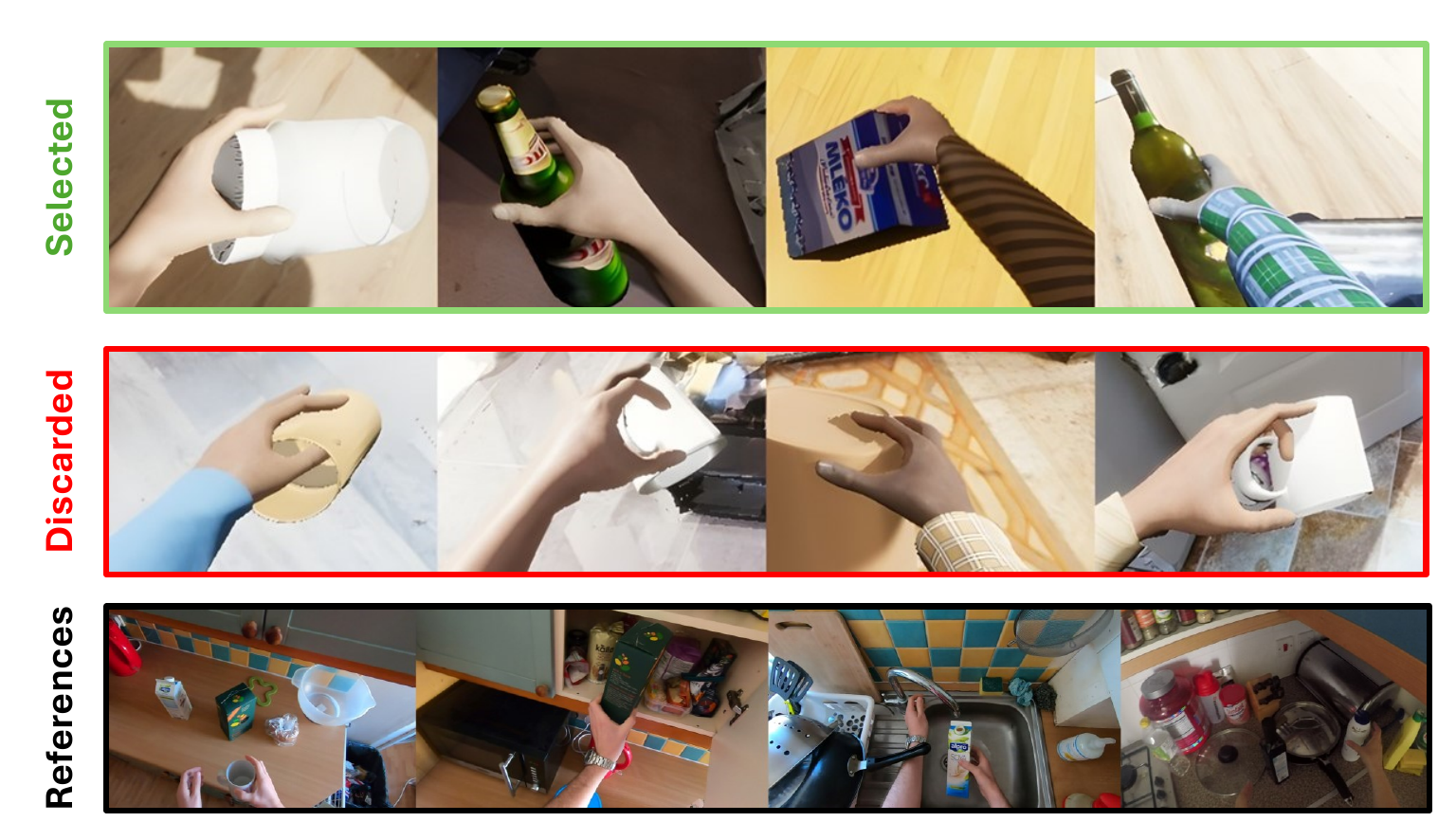}
    \caption{The green row shows examples of selected grasps aligned with real grasps from \textit{VISOR} (black row). The red row contains discarded cases due to misalignment or unrealistic grasp configurations.}
    \label{fig:grasps_aligned}
\end{figure*}

\subsubsection{Aligning Grasps}
To generate a synthetic dataset containing only hand grasps that closely align with real-world hand poses, we adopted the following approach.
First, we extracted 2D hand keypoints from \textit{VISOR} training images using \textit{MMPose}~\footnote{We used the following algorithm for 2D hand keypoints: \url{https://github.com/open-mmlab/mmpose/tree/main/configs/hand_2d_keypoint}}~\cite{mmpose2020}. These keypoints served as the foundation for identifying realistic hand-object interactions.
Next, for each virtual object in \textit{DexGraspNet}~\cite{wang2023dexgraspnet}, we generated a diverse set of images, with each object being represented by approximately 150 to 300 distinct grasps, ensuring a wide range of potential interactions. To ensure the synthetic grasps matched real-world interactions, we compared them to the \textit{VISOR} extracted keypoints using Euclidean distance as a similarity metric. This allowed us to select and retain only those grasps most similar to the real-world examples. Finally, using the selected grasps, we generated a synthetic dataset of 5,000 images, each containing one of the chosen synthetic grasps.

The results in Table~\ref{tab:target_visor} demonstrate that including target grasps enhances model performance. The \textit{Overall} mAP increases from $31.52\%$ to $31.84\%$, showing improved generalization. Notably, the \textit{H} metric rises from $79.58\%$ to $82.01\%$, highlighting better detection of hands. The \textit{H+C} metric also improves from $30.19\%$ to $31.95\%$, indicating enhanced hand-object interaction recognition. The \textit{O} metric increases from $7.35\%$ to $7.97\%$, demonstrating that target grasps effectively boost object recognition. Overall, these findings highlight the value of aligning synthetic grasps with real-world examples. See Figure~\ref{fig:grasps_aligned} for examples of selected aligned hand grasps. 

\begin{figure*}[t]
    \centering
    \includegraphics[width=\linewidth]{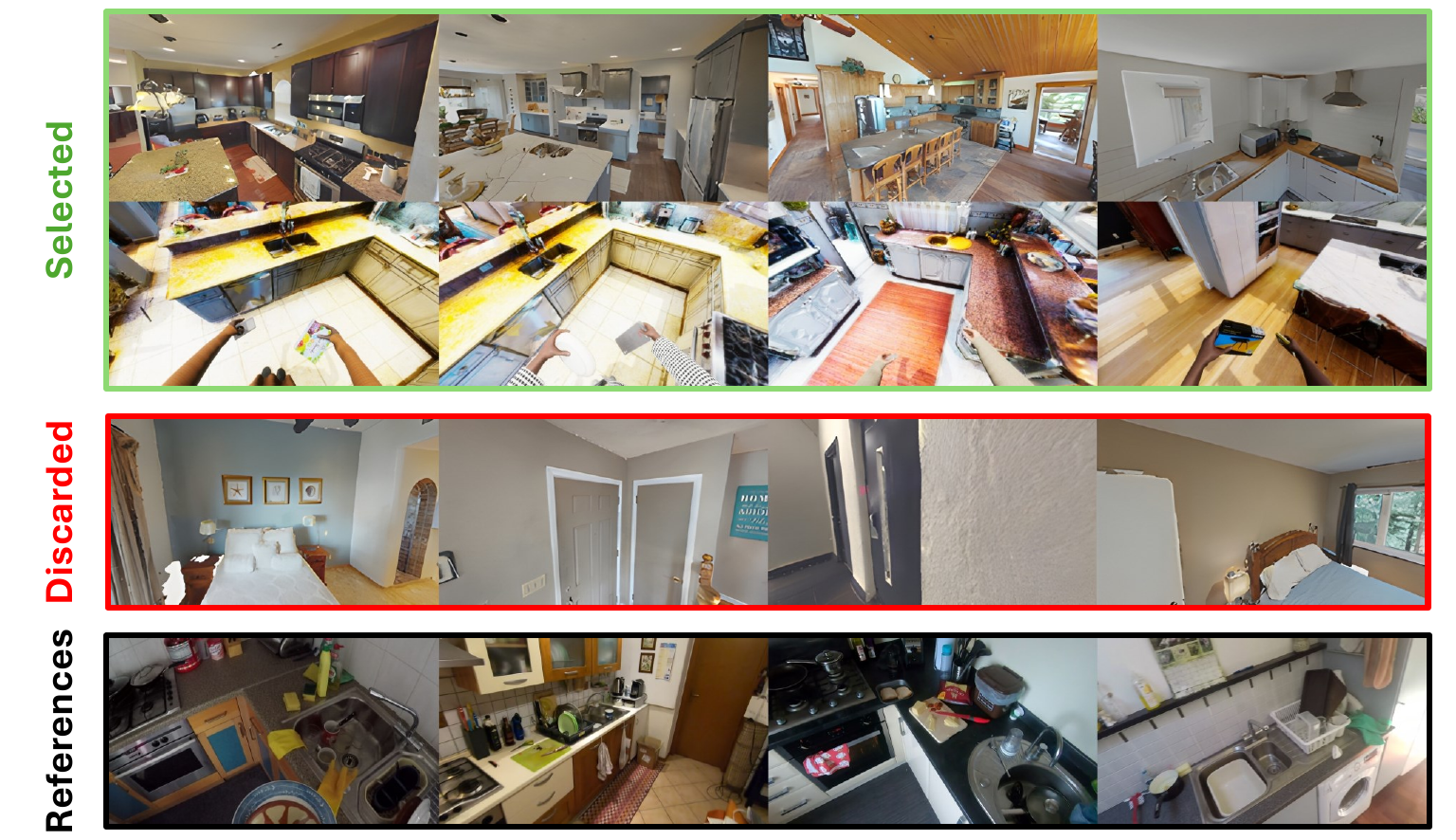}
    \caption{Green rows contain some examples of selected synthetic images aligned with \textit{VISOR} images (black row). Note that the selected environments mainly present kitchens. The red row shows discarded examples due to misalignment or unsuitable viewpoints.}
    \label{fig:envs_aligned}
\end{figure*}

\subsubsection{Aligning Environments}\label{sec:envs_visor}
In this experiment, we aimed to reduce the domain gap between synthetic and real-world environments. Aligning synthetic environments with real-world ones is crucial for improving the model's robustness across diverse real-world scenarios. To achieve this, we adopted a feature alignment strategy similar to the one applied for objects (Section~\ref{sec:object_align}). Specifically, we generated a set of synthetic views by randomly sampling positions for each environment in the HM3D~\cite{ramakrishnan2021habitat} dataset. For real-world environments, to maintain the focus on the scenes rather than hand-object interactions, we extracted frames from the \textit{VISOR} training videos at a rate of 1 frame per second, trained a hand detector using only synthetic data, and selected only frames where no hands were present. Once the real environment views were selected, we applied \textit{DBSCAN} clustering to filter out outliers. Then, we used the cosine distance between the features of the synthetic views and the mean feature vectors of the real-world clusters to identify the most similar synthetic views. Finally, we generated a dataset of 5,000 hand-object interaction images, with 70\% of the images captured from the selected viewpoints and 30\% from randomly sampled positions to introduce additional variability. Figure~\ref{fig:envs_aligned} shows some examples of synthetic images in the selected viewpoints. 
The results in Table~\ref{tab:target_visor} show the impact of aligning synthetic and real-world environments on model performance. While there is a marginal improvement in the \textit{Overall} mAP, from $31.52\%$ to $31.59\%$, the hand metrics (\textit{H} and \textit{H+C}) improve more noticeably. Specifically, the \textit{H} metric increases from $79.58\%$ to $81.22\%$, and the \textit{H+C} metric from $30.19\%$ to $32.68\%$. However, the performance on the object metric (\textit{O}) slightly decreases from $7.35\%$ to $7.16\%$, suggesting that the focus on environments may have led to learned representations that prioritize contextual information over individual object detection, negatively impacting the object-specific metric.

\subsubsection{Combining Objects, Grasps, and Environments}
In this experiment, we combined the previous strategies for aligning the objects, grasps, and environments. The goal was to evaluate how combining these three aspects affects overall model performance when used collectively rather than individually. Table~\ref{tab:target_visor} shows that combining all three strategies boosts the performance compared to the baseline approach for all the considered metrics. Specifically, for the set with 5k images, the \textit{Overall} mAP increases from $31.52\%$ (Baseline) to $32.59\%$. The results reveal that object alignment has the most significant impact on performance, as the improvements from the other strategies are minimal in comparison. This suggests that aligning the object alone may be sufficient to achieve similar performance improvements, highlighting the importance of object alignment in the process.

Considering the set with 30k training images, the ``Target All'' approach significantly enhances the model's performance, resulting in an overall mAP increase from $33.33\%$ (Baseline) to $33.98\%$. Focusing on the other breakdown metrics, the \textit{H} metric rises from $80.16\%$ to $83.20\%$, the \textit{H+C} metric increases from $33.47\%$ to $34.19\%$, and the \textit{O} metric rises from $8.35\%$ to $9.22\%$. These improvements further highlight the effectiveness of aligning objects, grasps, and environments to boost the model's performance.

\begin{table*}[!t]
	\centering
	\caption{Results on the \textit{EgoHOS}~\cite{EgoHos_jianbo_eccv22} test set.} 
	  
	\begin{minipage}{\linewidth} 
		\centering
		\textbf{a) Unsupervised Setting} \\
		  
		\resizebox{\linewidth}{!}{
			\begin{tabular}{c|c|c|cccc}
				\hline
				\textbf{\% Real Labeled Data} & \textbf{Approach} & \textbf{Overall} & H & H+S & H+C & O \\
				\hline
				\multirow{2}{*}{0\%}                     & \cellcolor{gray!30} 
				Synthetic-Only    & \cellcolor{gray!30} 07.16  & \cellcolor{gray!30} 18.25                           & \cellcolor{gray!30} 15.93                           & \cellcolor{gray!30} 05.33                           & \cellcolor{gray!30} 01.24                                                      \\ 
				&  UDA    
				&  \textbf{28.16}
				&  \textbf{70.30}             &  \textbf{59.21}             &  \textbf{20.84}             &  \textbf{09.65}                         \\ 
				\hline
				\multicolumn{2}{c|}{Absolute Improvement}  & $\textcolor{blue}{\textbf{+21.00}}$ & $+52.05$ & $+43.28$ & $+15.51$ & $+8.41$  \\ 
				\hline
			\end{tabular}
		}
	\end{minipage}
	\vspace{1em}
	      
	\begin{minipage}{\linewidth} 
		\centering
		\textbf{b) Semi-supervised Setting} \\
		  
		\resizebox{\linewidth}{!}{
			\begin{tabular}{c|c|c|cccc}
				
				\hline
				\textbf{\% Real Labeled Data}  \\
				\hline

				\multirow{3}{*}{\makecell{10\% \\ (857 images)}}          	& \cellcolor{gray!30} Real-Only   
				& \cellcolor{gray!30} 28.44   
				& \cellcolor{gray!30} 76.28                           & \cellcolor{gray!30} 68.92                           & \cellcolor{gray!30} 35.84                           & \cellcolor{gray!30} 16.59                                                   \\ 
				&  Synthetic+Real 
				&  28.74
				&  77.15                      &  71.64                      &  39.25                      &  17.33                                             \\ 
				&  SSDA 
				&  \textbf{\underline{36.68}}
				&  \textbf{\underline{83.25}} &  \textbf{\underline{73.72}} &  \textbf{\underline{47.20}} &   \textbf{\underline{22.40}} \\ 
				\hline
				\multicolumn{2}{c|}{Absolute Improvement}  
				& $\textcolor{blue}{\textbf{+8.24}}$
				& $+6.97$ &  $+4.80$   & $+11.36$  & $+5.81$  \\
							
				\hline

				\multirow{3}{*}{\makecell{25\% \\ (2,026 images)}}         & \cellcolor{gray!30} Real-Only       & \cellcolor{gray!30} 33.73     & \cellcolor{gray!30} 78.94                           & \cellcolor{gray!30} 70.62                           & \cellcolor{gray!30} 41.67                           & \cellcolor{gray!30} 21.83                                                     \\ 
				&  Synthetic+Real
				&  33.78
				&  79.60                      &  71.61                      &  46.11                      &  19.87                                            \\ 
				&  SSDA     
				&  \textbf{\underline{37.16}}
				&  \textbf{\underline{83.79}} &  \textbf{\underline{74.28}} &  \textbf{\underline{49.00}} &   \textbf{\underline{23.82}}  \\ 
				\hline
				\multicolumn{2}{l|}{Absolute Improvement}  
				& $\textcolor{blue}{\textbf{+3.43}}$
				& $+4.85$ &  $+3.66$   & $+7.33$  & $+1.99$ \\

				\hline
				\multirow{3}{*}{\makecell{50\% \\ (4,379 images)}}  		& \cellcolor{gray!30} Real-Only		
				& \cellcolor{gray!30} 36.30  & \cellcolor{gray!30} 81.82                           & \cellcolor{gray!30} 73.63                           & \cellcolor{gray!30} 47.27                           & \cellcolor{gray!30} 25.73                                                   \\ 
				&  Synthetic+Real
				&  34.30 
				&  82.54                      &  74.03                      &  47.92                      &  23.47                                           \\
				&  SSDA 
				&  \textbf{\underline{39.85}}	
				&  \textbf{\underline{85.17}}						&  \textbf{\underline{76.80}}						&  \textbf{\underline{52.58}}						&  \textbf{\underline{26.90}}											\\ 
				\hline
				\multicolumn{2}{c|}{Absolute Improvement}  
				& $\textcolor{blue}{\textbf{+3.55}}$
				& $+3.97$ &  $+3.17$   & $+5.31$  & $+1.17$ \\
				   
				\hline
 
			\end{tabular}
		}
	\end{minipage}
	\vspace{1em}
	       
	\begin{minipage}{\linewidth} 
		\centering
		\textbf{c) Fully-supervised Setting} \\
		  
		\resizebox{\linewidth}{!}{
			\begin{tabular}{c|c|c|cccc}
				\hline
				\textbf{\% Real Labeled Data}   \\
				\hline
				\multirow{3}{*}{\makecell{100\% \\ (8,758 images)}}        & \cellcolor{gray!30} Real-Only    & \cellcolor{gray!30} 36.16        & \cellcolor{gray!30} 84.39                           & \cellcolor{gray!30} 76.24                           & \cellcolor{gray!30} 51.81                           & \cellcolor{gray!30} 26.46                                                     \\ 
				&  Synthetic+Real
				&  34.68
				&  84.56                      &  71.56                      &  49.72                      &  23.16                                            \\
				&  FSDA    
				&  \textbf{\underline{39.61}}
				&  \textbf{\underline{85.58}} &  \textbf{\underline{76.80}} &  \textbf{\underline{51.99}} &  \textbf{\underline{27.05}}  \\
				\hline
				\multicolumn{2}{c|}{Absolute Improvement}
				& $\textcolor{blue}{\textbf{+3.45}}$
				& $+1.19$  & $+0.56$  & $+0.18$  & $+0.59$   \\

				\hline
			\end{tabular}
		}
	\end{minipage}
	\label{tab:hos_egohos}
\end{table*}

\subsection{Results on EgoHOS} \label{sec:egohos_results}
Table~\ref{tab:hos_egohos} shows the results obtained on the test set of \textit{EgoHOS}~\cite{EgoHos_jianbo_eccv22}. As in previous experiments, training exclusively on synthetic data (as shown in Table~\ref{tab:hos_egohos}-a) leads to poor performance. Specifically, the \textit{Synthetic-Only} approach achieves an Overall AP of $7.16\%$, which is approximately a $27\%$ lower compared to the fully supervised baseline trained on $100\%$ labeled real data (Table~\ref{tab:hos_egohos}-c). The \textit{UDA} approach shows significant improvements over the \textit{Synthetic-Only} baseline, boosting the Overall Mask AP by $+21.00\%$, and achieving gains of $+52.05\%$, $+43.28\%$, $+15.51\%$, and $+8.41\%$ in the \textit{H}, \textit{H+S}, \textit{H+C}, and \textit{O} metrics, respectively.

In the semi-supervised setting (Table~\ref{tab:hos_egohos}-b), both \textit{Synthetic+Real} and \textit{SSDA} outperform the \textit{Real-Only} baseline. \textit{SSDA} shows the highest improvements, surpassing \textit{Real-Only} by $+8.24\%$, $+3.43\%$, and $+3.55\%$ in Overall AP when using 10\%, 25\%, and 50\% labeled real data, respectively.

It is worth noting that the performance gain provided by synthetic data tends to decrease as the amount of real labeled data increases (e.g., from $+8.24\%$ at $10\%$ labels to $+3.55\%$ at $50\%$). This behavior is expected since, in low-data regimes, synthetic data serves as a crucial source of supervision by compensating for the scarcity of real annotations. As the amount of labeled real data increases, the \textit{Real-Only} baseline naturally achieves higher performance, which reduces the marginal benefit of domain adaptation and narrows the room for further improvement.

In the fully supervised setting (Table~\ref{tab:hos_egohos}-c), the \textit{FSDA} approach achieves the highest scores across all metrics, outperforming \textit{Real-Only} by $+3.45\%$ in Overall AP. It is worth noting that the \textit{SSDA} model trained with just 25\% of labeled real data (2,026 images) reached the performance of the \textit{Real-Only} model trained on 100\% of the real data (8,758 images), with a $+0.52\%$ improvement in Overall AP.

These findings highlight the effectiveness of synthetic data in minimizing the reliance on large amounts of real labeled data, while also improving performance compared to standard fully supervised approaches.

\begin{figure*}[!ht]
    \centering
    \includegraphics[width=\linewidth]{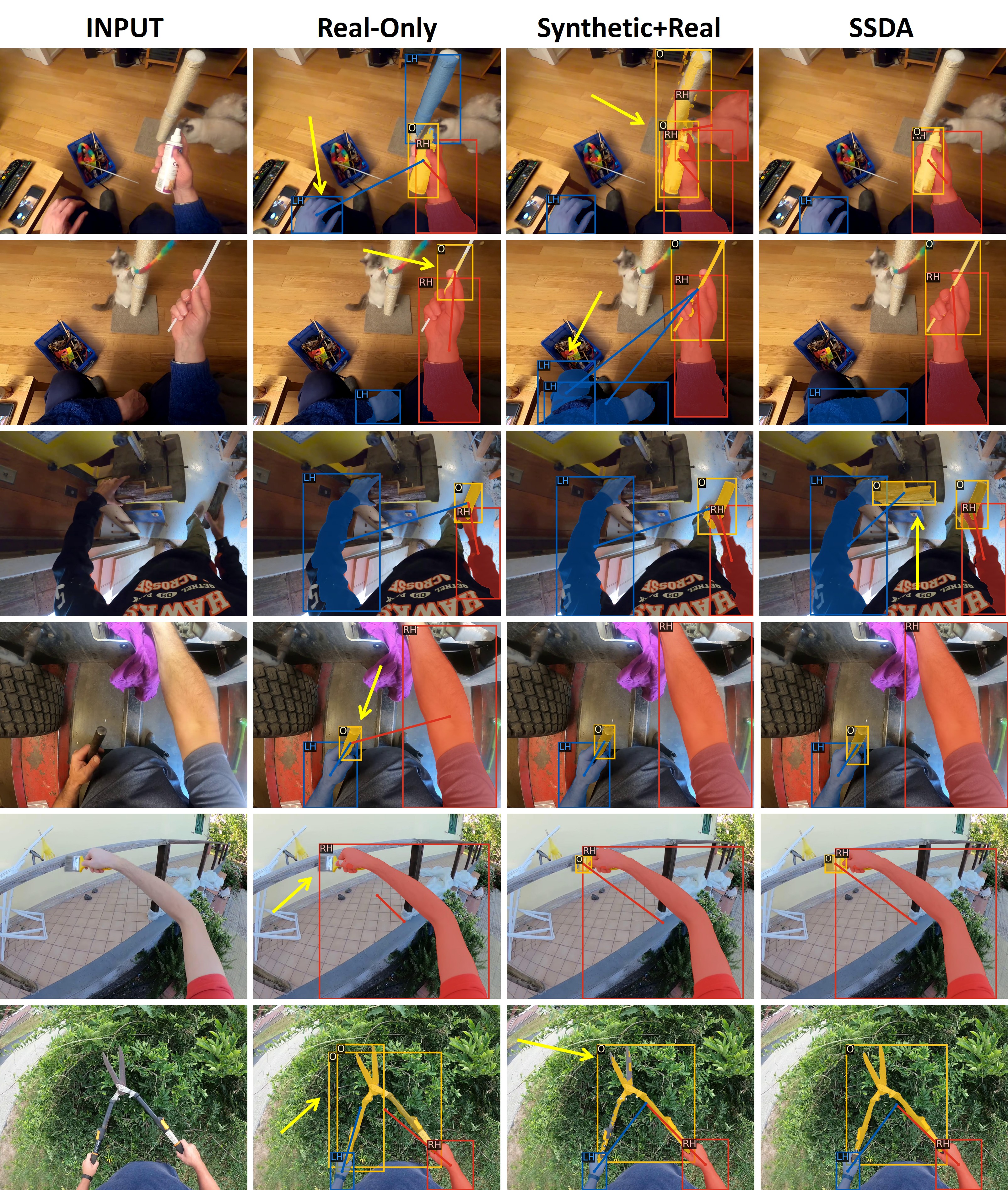}
    \caption{Qualitative examples on \textit{EgoHOS}.}
    \label{fig:qualitative_examples_egohos}
\end{figure*}

Figure~\ref{fig:qualitative_examples_egohos} shows qualitative examples on the \textit{EgoHOS} test set. The \textit{SSDA} approach consistently enhances segmentation quality and reduces false positives, as observed in the first, second, third, and last rows. In contrast, the \textit{Real-Only} method fails to detect the object being interacted with (the paintbrush) in the fifth row.

\clearpage

\begin{table*}[t]
	\centering
	\caption{Results on the \textit{ENIGMA-51}~\cite{ragusa2024enigma} test set.} 
	\begin{minipage}{\linewidth} 
		\centering
		\textbf{a) Unsupervised Setting}
		\resizebox{\linewidth}{!}{
			                                                    
			\begin{tabular}{c|c|c|c|cccc}
				\hline
				\textbf{\% Real Labeled Data} & \textbf{Approach}                    & \textbf{In-domain}                    & \textbf{Overall}                                      & H                                                     & H+S                                                   & H+C                                                   & O                                                     \\
				\hline
				\multirow{4}{*}{0\%}   
				                
				                              & \cellcolor{gray!30} Synthetic-Only & \cellcolor{gray!30}                 & \cellcolor{gray!30} 00.21                           & \cellcolor{gray!30} 01.07                           & \cellcolor{gray!30} 00.11                           & \cellcolor{gray!30} 00.03                           & \cellcolor{gray!30} 00.99                           \\

				                              & \cellcolor{gray!30} Synthetic-Only & \cellcolor{gray!30} \checkmark      & \cellcolor{gray!30} 12.85                           & \cellcolor{gray!30} 56.05                           & \cellcolor{gray!30} 35.14                           & \cellcolor{gray!30} 15.24                           & \cellcolor{gray!30} 4.79                            \\

				                              &  UDA       &             &  6.87                       &  42.81                      &  14.52                      &  7.97                       &  3.29                       \\

				                              &  UDA       &  \checkmark &  \textbf{\underline{34.78}} &  \textbf{\underline{78.83}} &  \textbf{\underline{70.91}} &  \textbf{\underline{28.14}} &  \textbf{\underline{25.84}} \\ \hline
				                  
				\multicolumn{3}{c|}{Absolute Improvement}   & $\textcolor{blue}{\textbf{+21.93}}$       & $+22.78$  & $+35.77$  & $+12.90$  & $+21.05$  \\
				\hline
			\end{tabular} 
		}
	\end{minipage}
	        
	\vspace{1em}
	        
	\begin{minipage}{\linewidth} 
		\centering
		\textbf{b) Semi-supervised Setting} \\
		\resizebox{\linewidth}{!}{
			       
			\begin{tabular}{c|c|c|c|cccc}
				            
				\hline
				\textbf{\% Real Labeled Data} & \textbf{Approach}               & \textbf{In-domain}                    & \textbf{Overall}                                      & H                                                     & H+S                                                   & H+C                                                   & O                                                     \\
				            
				\hline
				\multirow{3}{*}{\makecell{10\% \\ (347 images)}}  
				            
				                              & \cellcolor{gray!30} Real-Only & \cellcolor{gray!30} \checkmark      & \cellcolor{gray!30} 45.39                           & \cellcolor{gray!30} 81.25                           & \cellcolor{gray!30} 76.22                           & \cellcolor{gray!30} 37.96                           & \cellcolor{gray!30} 39.53                           \\

				                              &  SSDA &             &  \textbf{\underline{57.08}} &  \textbf{\underline{85.40}} &  \textbf{\underline{78.62}} &  \textbf{\underline{43.56}} &  \textbf{\underline{46.97}} \\

				                              &  SSDA &  \checkmark &  56.69                      &  84.58                      &  78.42                      &  41.17                      &  46.50                      \\ 
				\hline
				
				\multicolumn{3}{c|}{Absolute Improvement}  & $\textcolor{blue}{\textbf{+11.69}}$ & $+4.15$ & $+2.40$ & $+5.60$ & $+7.44$  \\
				            
				\hline
				
				         
				\multirow{3}{*}{\makecell{25\% \\ (870 images)}}  
				            
				                              & \cellcolor{gray!30} Real-Only & \cellcolor{gray!30} \checkmark      & \cellcolor{gray!30} 51.83                           & \cellcolor{gray!30} 82.95                           & \cellcolor{gray!30} 78.70                           & \cellcolor{gray!30} 43.52                           & \cellcolor{gray!30} 45.25                           \\ 
				            
				                              &  SSDA &             &  58.17                      &  \textbf{\underline{84.99}} &  \textbf{\underline{80.41}} &  \textbf{\underline{46.31}} &  49.34                      \\

				&  SSDA   &  \checkmark &  \textbf{\underline{59.48}}        
				&  84.85 &  80.30                      &  44.24                      &  \textbf{\underline{\textbf{49.37}}}  \\

				\hline
				\multicolumn{3}{c|}{Absolute Improvement} & $\textcolor{blue}{\textbf{+7.65}}$  & $+2.04$ & $+1.71$ & $+2.79$ & $+4.12$ \\
				\hline

				
				\multirow{3}{*}{\makecell{50\% \\ (1,739 images)}}  
				& \cellcolor{gray!30} Real-Only & \cellcolor{gray!30} \checkmark   & \cellcolor{gray!30} 57.62
				& \cellcolor{gray!30} 84.65                           & \cellcolor{gray!30} 80.43                  & \cellcolor{gray!30} 47.41                           & \cellcolor{gray!30} 48.79                                                     \\

				                              &  SSDA &             &  \textbf{\underline{63.25}} &  \textbf{\underline{85.67}} &  82.00                      &  \textbf{\underline{52.20}} &  \textbf{\underline{52.56}} \\

				                              &  SSDA &  \checkmark &  61.93                      &  85.12                      &  \textbf{\underline{82.01}} &  48.96                      &  51.94                      \\
				\hline
				                
				\multicolumn{3}{c|}{Absolute Improvement} & $\textcolor{blue}{\textbf{+5.63}}$  & $+1.02$ & $+1.58$ & $+4.79$ & $+3.77$  \\                   
				\hline  
			\end{tabular}
		}
	\end{minipage}
	
	\vspace{1em}
	       
	\begin{minipage}{\linewidth} 
		\centering
		\textbf{c) Fully-supervised Setting} \\
		\resizebox{\linewidth}{!}{
			       
			\begin{tabular}{c|c|c|c|cccc}
				    
				\hline
				\textbf{\% Real Labeled Data} & \textbf{Approach}               & \textbf{In-domain}               & \textbf{Overall}                                      & H                                                     & H+S                                                   & H+C                                                   & O                                \\
				\hline
				\multirow{3}{*}{\makecell{100\% \\ (3,479 images)}}
				                              & \cellcolor{gray!30} Real-Only & \cellcolor{gray!30} \checkmark & \cellcolor{gray!30} 63.84                           & \cellcolor{gray!30} 85.01                           & \cellcolor{gray!30} 81.05                           & \cellcolor{gray!30} 52.32                           & \cellcolor{gray!30} 51.35      \\

				                              &  FSDA &        &  \textbf{\underline{64.41}} &  \textbf{\underline{85.94}} &  \textbf{\underline{82.91}} &  \textbf{\underline{54.13}} &  52.50 \\

				&  FSDA   &  \checkmark &  64.20       
				&  85.37          &  82.45                      &  51.60          &  \underline{\textbf{53.30}}                       \\
				            
				\hline
				\multicolumn{3}{c|}{Absolute Improvement} & $\textcolor{blue}{\textbf{+0.57}}$ & $+0.93$  & $+1.86$ & $+1.81$ & $+1.95$  \\
				            
				\hline
				    			      
			\end{tabular}
		}
	\end{minipage}
	
	\label{tab:hos_enigma}
\end{table*}

\subsection{Results on ENIGMA-51}\label{sec:enigma_results}
Table~\ref{tab:hos_enigma} shows the results on \textit{ENIGMA-51}~\cite{ragusa2024enigma}, comparing model performance when using both in-domain and out-domain synthetic data (see Section~\ref{tab:datasets} and Figure~\ref{fig:in-domain}).

In the unsupervised setting (Table~\ref{tab:hos_enigma}-a), training exclusively on out-domain synthetic data leads to poor results, reinforcing the evidence of the domain gap between synthetic and real-world data. In contrast, using in-domain synthetic data narrows this gap significantly, with the \textit{Overall} AP increasing from $0.21\%$ to $12.85\%$. Using \textit{UDA} with synthetic data enhances performance further, particularly with in-domain data, which provides a boost of $+21.93\%$ in \textit{Overall} AP and significant improvements in breakdown metrics: $+22.78\%$ \textit{H}, $+35.77\%$ \textit{H+S}, $+12.90\%$ \textit{H+C}, and $+21.05\%$ \textit{O}. The type of synthetic data proves to be a crucial factor in unsupervised domain adaptation. Models trained with in-domain synthetic data surpass those trained with out-domain data by a significant $27.91\%$ in \textit{Overall} AP, demonstrating the importance of domain-specific synthetic data in this setting.

\begin{figure*}[!ht]
    \centering
    \includegraphics[width=\linewidth]{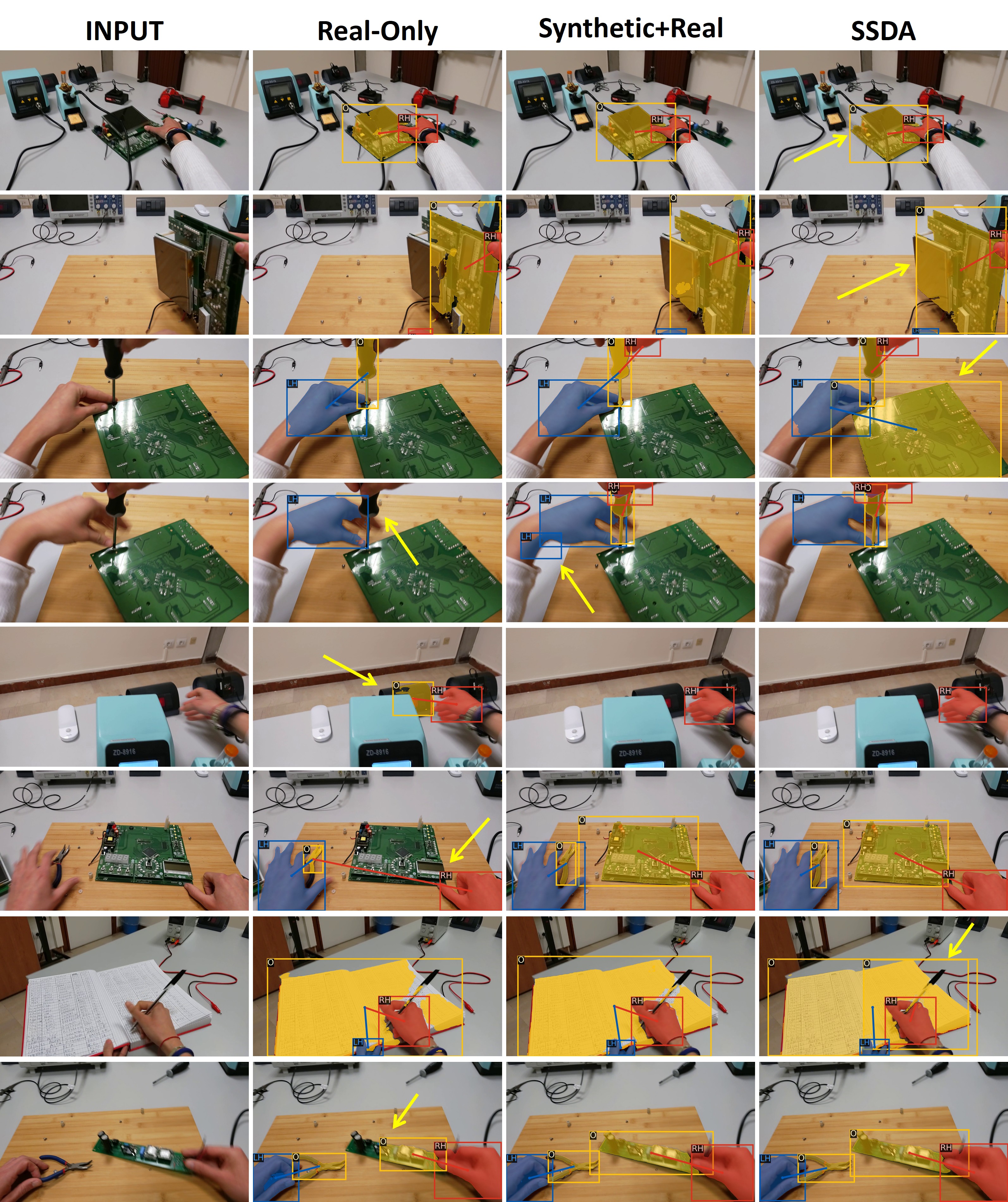}
    \caption{Qualitative examples on \textit{ENIGMA-51}.}
    \label{fig:qualitative_examples_enigma}
\end{figure*}
\clearpage

In the semi-supervised setting (Table~\ref{tab:hos_enigma}-b), \textit{SSDA} offers consistent improvements over the baseline, with gains of $+11.69\%$, $+7.65\%$, and $+5.63\%$ in \textit{Overall} AP when using 10\%, 25\%, and 50\% labeled real data, respectively. As the amount of real supervision increases, the contribution of synthetic data gradually decreases, as also observed in Sections~\ref{sec:visor_results} and~\ref{sec:egohos_results}. Accordingly, in the fully supervised setting (Table~\ref{tab:hos_enigma}-c), the results improve only slightly, with an \textit{Overall} AP increase of $+0.57\%$. Interestingly, the difference between in-domain and out-domain synthetic data becomes less pronounced in semi- and fully-supervised settings, with both yielding comparable results. This indicates that even small amounts of labeled real data can mitigate the domain gap, making the choice of synthetic data source less critical as real data is introduced.

Figure~\ref{fig:qualitative_examples_enigma} shows qualitative examples on the \textit{ENIGMA-51} test set. The \textit{SSDA} approach significantly improves segmentation across all examples, particularly in the third, fourth, fifth, sixth, and seventh rows, while minimizing false positives when compared to the \textit{Real-Only} approach.

\subsubsection{Targeting Field Of View on ENIGMA-51}
Table~\ref{tab:fov_enigma} shows the results of aligning the Field of View (FOV) of out-domain synthetic data with that of \textit{ENIGMA-51}. To achieve this alignment, we first identified the specifications of the acquisition device used in the \textit{ENIGMA-51} dataset (HoloLens 2). We replicated the camera parameters within our simulation environment, ensuring that the simulated camera matched its characteristics. This approach allowed us to generate synthetic images that maintained the same field of view as those from \textit{ENIGMA-51}. Through this alignment, we aimed to evaluate whether adjusting the FOV could improve method performance. The results indicate that modifications to the FOV can lead to a modest improvement in performance. However, it is evident that, despite these adjustments, the performance gap between out-domain synthetic data and in-domain data persists. This further highlights the importance of using in-domain synthetic data to achieve optimal performance within a specific domain.

\begin{table*}[t]
    \centering
    \caption{Performance comparison when aligning the FOV of out-domain synthetic data to that of \textit{ENIGMA-51}. }
    \resizebox{\linewidth}{!}{
        \begin{tabular}{lcc|c|ccc}
            \hline
            \textbf{Approach} & \textbf{Target FOV} &\textbf{In-domain} & \textbf{Overall} &  H              & H+C            & O              \\
            \hline               
            Synthetic-Only    &  &                            & 00.21 & 01.07 & 00.03 & 00.99           \\
            Synthetic-Only    & \checkmark &                  & 05.67             & 15.78          & 02.66           & 02.31           \\
            Synthetic-Only    &\checkmark& \checkmark         & \textbf{12.85}   & \textbf{56.05} & \textbf{15.24} & \textbf{4.79}  \\ \hline
        \end{tabular}
    }
    \label{tab:fov_enigma}
\end{table*}

\subsection{Importance of Accurate Simulation of Hand-Object Interactions}
In this section, we analyze how accurate simulation of hand-object interactions contributes to building robust and reliable systems for HOI detection in complex real-world scenarios. Specifically, we first investigate the importance of background environment coherence in improving detection performance (Section~\ref{sec:background_coherence}). Next, we conduct a detailed evaluation focusing on the prediction of hand side and contact, which are critical components for accurately understanding hand-object interactions (Section~\ref{sec:eval_hand_contact_side}). Finally, we compare the efficacy of simulation-based data generation with traditional augmentation techniques to determine which approach offers superior performance for training robust models (Section~\ref{sec:aug_vs_simulation}).

\subsubsection{Evaluating the Importance of the Environment Background in Hand-Object Interaction Detection}\label{sec:background_coherence}
In this experiment, we aimed to investigate the impact of environment background coherence on the performance of our hand-object interaction detection model. To assess this, we utilised our baseline synthetic dataset and created two variants by replacing only the backgrounds of the images: one with backgrounds sourced from \textit{ImageNet}~\cite{russakovsky2015imagenet} and another with backgrounds taken from the \textit{VISOR} dataset (see Section~\ref{sec:envs_visor}).
The results in Table~\ref{tab:background_imagenet_visor} show the significant influence of background choice on the performance of the model. Training with the baseline dataset yields the highest overall mAP of 31.52\%, demonstrating the importance of maintaining a coherent background. In contrast, using \textit{ImageNet} and \textit{VISOR} backgrounds leads to a notable drop in performance, with \textit{Overall} mAPs of 28.86\% and 27.41\%, respectively. While the hand metric (\textit{H}) improves with \textit{VISOR} backgrounds, reaching 82.30\%, both the \textit{H+C} and \textit{O} metrics decrease considerably. Specifically, the \textit{H+C} metric drops to 6.09\% (\textit{VISOR}) and 27.16\% (\textit{ImageNet}), and the \textit{O} metric falls to 2.56\% (\textit{VISOR}) and 5.47\% (\textit{ImageNet}). These results highlight that even with identical HOIs, the background environment significantly influences model accuracy, supporting the need for background coherence to improve hand-object interaction detection.

\begin{table*}[t]
	\centering
	\caption{Performance comparison on the \textit{VISOR} validation dataset using the UDA model trained on 5k synthetic images with different backgrounds. The baseline with coherent backgrounds outperforms the models trained with \textit{ImageNet} and \textit{VISOR} backgrounds, highlighting the importance of background alignment for cross-domain performance.}
	\resizebox{0.7\linewidth}{!}{
		\begin{tabular}{l|c|ccc}
			\hline
			\textbf{Background}   & \textbf{Overall} & H              & H+C            & O    \\
			\hline            
			Baseline              & \textbf{31.52}     & 79.58          & \textbf{30.19}     & \textbf{7.35} \\
			ImageNet              & 28.86	           & 77.38	        & 27.16              & 5.47 \\
			VISOR                 & 27.41	           & \textbf{82.30}	& 6.09	             & 2.56 \\
			\hline 
		\end{tabular}
	}
	\label{tab:background_imagenet_visor}
\end{table*}

\subsubsection{Evaluating Hand Contact and Side Predictions}\label{sec:eval_hand_contact_side}
To focus on the specific downstream tasks, we computed \textit{F1 scores} for hand contact and side predictions, independent of hand segmentation accuracy. As shown in Table~\ref{tab:f1_score}, the integration of synthetic data leads to consistent improvements, highlighting the value of simulating interactions beyond enhancing semantic segmentation. On the \textit{VISOR} dataset, the \textit{Contact F1 score} increased from $79.46$\% to $81.65$\%, with the \textit{Side F1 score} improving from $98.92$\% to $99.06$\%. A similar trend is observed on the \textit{EgoHOS} dataset, where \textit{Contact F1} increased from $78.78$\% to $79.25$\%, and \textit{Side F1} increased from $98.33$\% to $98.87$\%. In the case of \textit{ENIGMA-51}, \textit{Contact F1} improved from $77.28$\% to $77.83$\%, and Side F1 from $98.80$\% to $99.11$\%. These results demonstrate the robustness and effectiveness of synthetic data in enhancing model performance for hand contact and side predictions across different datasets.

\begin{table*}[t]
    \centering
    \caption{Comparison of FSDA vs Real-Only for F1 Scores of Contact and Side.}
    \resizebox{0.8\linewidth}{!}{
    		\begin{tabular}{lccc}
    			\hline
    			\textbf{Dataset} & \textbf{Approach} & \textbf{Contact F1} & \textbf{Side F1} \\
    			\hline               
    			VISOR            & Real-Only         &  79.46           &   98.92       \\
    			VISOR            & FSDA              &  \textbf{81.65}           &  \textbf{99.06}        \\ 
    			\hline 
    			EgoHOS          & Real-Only         &  78.78           & 98.33         \\ 
    			EgoHOS          & FSDA              &  \textbf{79.25}           & \textbf{98.87}          \\ 
    			\hline
    			ENIGMA-51        & Real-Only         &  77.28            & 98.80              \\ 
    			ENIGMA-51        & FSDA              &  \textbf{77.83}             & \textbf{99.11}              \\     
    						            
    			\hline
    		\end{tabular}
    	}
         \label{tab:f1_score}
\end{table*}

\begin{figure*}[t]
    \centering
    \includegraphics[width=\linewidth]{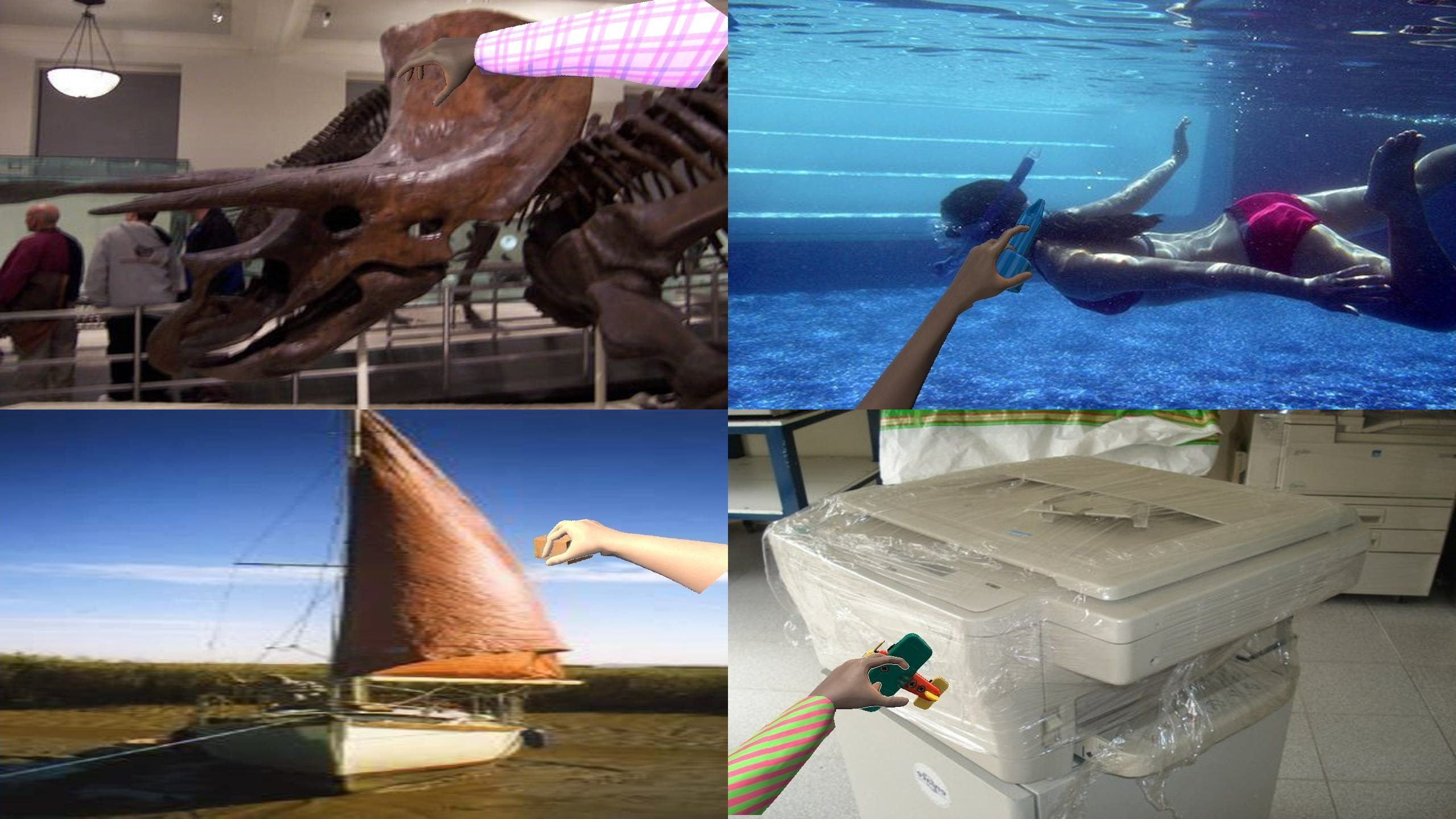}
    \caption{Simple HOI generation approach: Object models and hands are randomly projected onto ImageNet backgrounds.}
    \label{fig:sim_vs_aug}
\end{figure*}

\subsubsection{Assessing the Efficacy of Simulation Compared to Classic Augmentations}\label{sec:aug_vs_simulation}
To verify whether the performance gains are driven by the proposed simulation pipeline rather than merely by the use of synthetic data with domain adaptation, we conducted a comparison against a traditional augmentation baseline. Specifically, we generated a baseline dataset of 3,500 images by randomly projecting object and hand models onto background images sampled from ImageNet, ignoring physical constraints (see Figure~\ref{fig:sim_vs_aug}). We then compared this baseline against a subset of 3,500 images from our physically simulated dataset. Both sets were used to train the same UDA model on \textit{VISOR} under identical conditions. We aim to demonstrate that while domain adaptation bridges the photorealism gap, it requires physically consistent data to effectively learn interaction patterns. As shown in Table~\ref{tab:sim_vs_aug}, the simulation-based approach significantly outperforms the augmentation baseline, particularly on interaction-sensitive metrics such as \textit{H+C} and \textit{Contact F1}. These results confirm that the realistic fidelity provided by simulation is a critical factor that classic augmentation techniques cannot replicate, even when paired with powerful adaptation methods.

\begin{table*}[t]
     \centering
     \caption{UDA Results on \textit{VISOR}: Comparison of simulation-based image generation and classic synthetic augmentations.}

    \resizebox{\linewidth}{!}{
        \begin{tabular}{l|c|cccc}
            \hline
            \textbf{Generation approach} & \textbf{Overall} & H              & H+C            & O             & Contact F1     \\
            \hline               
            Synthetic augmentation       & 19.62            & 62.40          & 11.91          & 3.60          & 8.78           \\
            Simulation (Ours)            & \textbf{30.54}   & \textbf{79.05} & \textbf{31.51} & \textbf{7.33} & \textbf{22.75} \\ 
                                                                        
            \hline
        \end{tabular}
    }
    \label{tab:sim_vs_aug}

\end{table*}

\section{Discussion}\label{sec:discussion}
This study was driven by eight research questions that were aimed at systematically assessing the role of synthetic data in egocentric HOI detection. We now revisit these questions and discuss the corresponding experimental findings. \\

\noindent\textbf{\textit{Q1-3: Is there a gap between synthetic and real data? What are its causes and how can it be reduced?}} \\
Despite recent advancements in simulation, a significant performance gap persists between models trained on synthetic versus real data. \\
\textbf{1) Magnitude:} Our experiments estimate this gap to range between $30\%$ and $40\%$ AP depending on the dataset. For instance, on \textit{VISOR}, the gap starts at $35.45\%$ (Real-Only vs. Synthetic-Only). \\
\textbf{2) Causes:} We attribute this discrepancy to three main factors: limited photo-realism, the complexity of modeling realistic hand-object interactions, and the diversity of context-aware characteristics (backgrounds and object semantics), as evidenced by the performance drop when using out-of-domain synthetic data. \\
\textbf{3) Mitigation:} This gap can be effectively bridged through domain adaptation. On \textit{VISOR}, Unsupervised Domain Adaptation (UDA) narrows the gap to $12.00\%$, while Semi-Supervised Domain Adaptation (SSDA), leveraging just small amounts of real labels, further minimizes it to $1.11\%$. Similar trends are consistently observed across EgoHOS and ENIGMA-51.\\

\noindent\textbf{\textit{Q4: Can synthetic data fully replace real-world data?}} \\
Currently, synthetic data alone cannot fully replace real-world data for egocentric HOI detection. Our \textit{Synthetic-Only} baselines consistently yield suboptimal performance across all datasets due to the persistent domain shift. However, while they are not a standalone replacement, our results demonstrate that synthetic data are a critical \textit{complementary} resource that significantly boosts performance when real labels are scarce or when used in adaptation frameworks. \\

\noindent\textbf{\textit{Q5: Is it possible to leverage synthetic data when real-world data is unlabeled?}} \\
Yes, synthetic data are highly effective in this scenario. By employing unsupervised domain adaptation to transfer knowledge from labeled synthetic data to unlabeled real data, we observe substantial improvements over training on synthetic data alone. For instance, on \textit{VISOR}, our UDA approach yields a $+23.45\%$ gain over the \textit{Synthetic-Only} baseline, achieving an Overall AP of $33.33\%$. While this is still $\sim10\%$ lower than a fully supervised oracle, it demonstrates that synthetic data can successfully drive learning even without any real-world annotations.\\

\noindent\textbf{\textit{Q6: Can synthetic data improve performance when only a small amount of real-world labeled data is available?}} \\
When real labels are scarce, combining them with synthetic data via Semi-Supervised Domain Adaptation (SSDA) yields major efficiency gains. Our experiments show that models trained with just $10\%$ of real labeled data plus synthetic data achieve performance levels comparable to baselines trained on $100\%$ of real data (e.g., $44.22\%$ vs. $45.33\%$ on \textit{VISOR}). This highlights the impact that synthetic data can have on model effectiveness, especially in scenarios where labeled data is scarce. \\

\noindent\textbf{\textit{Q7: What scale of synthetic data is needed?}} \\
Our analysis indicates that models benefit significantly from large volumes of synthetic data up to a saturation point. On \textit{VISOR}, performance plateaus between 22,000 and 30,000 synthetic images. Extending the scale further to 80,000 images yields negligible gains, suggesting that $\sim$30k images is a cost-effective optimal scale for this task. \\

\noindent\textbf{\textit{Q8: Is in-domain synthetic data beneficial?}} \\
Yes, but the impact depends heavily on the supervision regime. \\
\textbf{1) Unsupervised Setting:} In-domain alignment is critical here. On the \textit{ENIGMA-51} dataset, in-domain synthetic data outperforms generic data by $\sim 10\%$ AP, enabling a significant boost to $34.78\%$ AP with UDA. Similarly, on \textit{VISOR}, our alignment strategies (objects/grasps/environments) yield consistent gains. \\
\textbf{2) Semi-Supervised Setting:} Interestingly, the necessity of strict alignment diminishes when real labels are available. Even small amounts of labeled real data act as a strong domain bridge, allowing models trained on generic synthetic data to achieve performance comparable to those trained on in-domain data. Thus, the effort of in-domain generation is most justified when real annotations are completely unavailable.

\subsection{Limitations and future work}
Our analysis is restricted to frame-level HOS detection and does not exploit temporal information. Accordingly, our simulator currently generates independent frames without enforcing temporal coherence. Extending the simulator to produce short temporally consistent sequences is an interesting direction for future work, especially for methods that explicitly model hand–object dynamics over time.
Furthermore, the current generation pipeline focuses on grasping interactions derived from \textit{DexGraspNet}. Consequently, complex object-object interactions (e.g., using a tool like a spatula to stir food) or cooperative bimanual actions (e.g., holding a large box with two hands) are not explicitly modeled. Future extensions could integrate physics-based simulation of tools and multi-hand grasping policies to cover these complex scenarios. \\

\section{Conclusion}\label{sec:conclusion}
In this work, we conducted a comprehensive investigation into the role of synthetic data for egocentric hand-object interaction detection. Through extensive experiments on three benchmarks, we showed that synthetic data, when combined with domain adaptation techniques, can significantly improve detection performance, particularly when real labeled data are scarce or unavailable. \\
Our findings indicate that, although a domain gap exists between synthetic and real data, it can be effectively reduced through unsupervised and semi-supervised domain adaptation approaches. Furthermore, we demonstrated that explicitly aligning objects, grasps, and environments to the target domain provides additional improvements, especially in fully unsupervised scenarios where no real labels are available.

We believe that our analysis provides valuable insights for future developments and applications in this area. Furthermore, the introduction of the \textit{HOI-Synth} benchmark and the associated data generation pipeline aims to support further research in this direction.

\section*{Acknowledgements}
This research has been supported by the project Future Artificial Intelligence Research (FAIR) – PNRR MUR Cod. PE0000013 - CUP: E63C22001940006. This research has been partially supported by the project EXTRA-EYE - PRIN 2022 - CUP E53D23008280006 - Finanziato dall'Unione Europea - Next Generation EU.

\bibliography{sn-bibliography}

\end{document}